\documentclass[10pt,twocolumn,letterpaper]{article}

\usepackage{iccv}
\usepackage{times}
\usepackage{epsfig}
\usepackage{graphicx}
\usepackage{amsmath}
\usepackage{amssymb}
\usepackage{booktabs}
\usepackage{multirow}
\usepackage{caption}
\usepackage{siunitx}
\usepackage{subfigure}
\usepackage{algorithm}
\usepackage{algpseudocode}
\usepackage{amsthm}
\usepackage{mathrsfs}
\usepackage{overpic}
\usepackage{color}

\usepackage{url}            
\usepackage{nicefrac}       
\usepackage{microtype}      

\usepackage{array}
\usepackage{cases}
\usepackage{booktabs}
\usepackage{multirow}
\usepackage{colortbl}

\usepackage[breaklinks=true,bookmarks=false]{hyperref}

\iccvfinalcopy 

\begin{document}

\title{Vulnerability of Appearance-based Gaze Estimation}

\author{Mingjie~Xu\textsuperscript{\rm 1}\qquad~Haofei~Wang\textsuperscript{\rm 2}\qquad~Yunfei~Liu\textsuperscript{\rm 1}\qquad~Feng~Lu\textsuperscript{\rm 1, 2, }\thanks{ Corresponding Author.}\\
	{\textsuperscript{\rm 1} State Key Laboratory of VR Technology and Systems, 
	School of CSE, Beihang University}  \\
	{\textsuperscript{\rm 2} Peng Cheng Laboratory, Shenzhen, China} \\
	\small{\texttt{xumingjies@gmail.com}} \qquad
	\small{\texttt{wanghf@pcl.ac.cn}} \qquad
	\small{\texttt{\{lyunfei,lufeng\}@buaa.edu.cn}}
}

\maketitle

\begin{abstract}

Appearance-based gaze estimation has achieved significant improvement by using deep learning. However, many deep learning-based methods suffer from the vulnerability property, i.e., perturbing the raw image using noise confuses the gaze estimation models. Although the perturbed image visually looks similar to the original image, the gaze estimation models output the wrong gaze direction. In this paper, we investigate the vulnerability of appearance-based gaze estimation. To our knowledge, this is the first time that the vulnerability of gaze estimation to be found. We systematically characterized the vulnerability property from multiple aspects, the pixel-based adversarial attack, the patch-based adversarial attack and the defense strategy. Our experimental results demonstrate that the CA-Net shows superior performance against attack among the four popular appearance-based gaze estimation networks, Full-Face, Gaze-Net, CA-Net and RT-GENE. This study draws the attention of researchers in the appearance-based gaze estimation community to defense from adversarial attacks.

\end{abstract}

\section{Introduction}

\begin{figure}[t]
   \begin{center}
         \includegraphics[width=\linewidth]{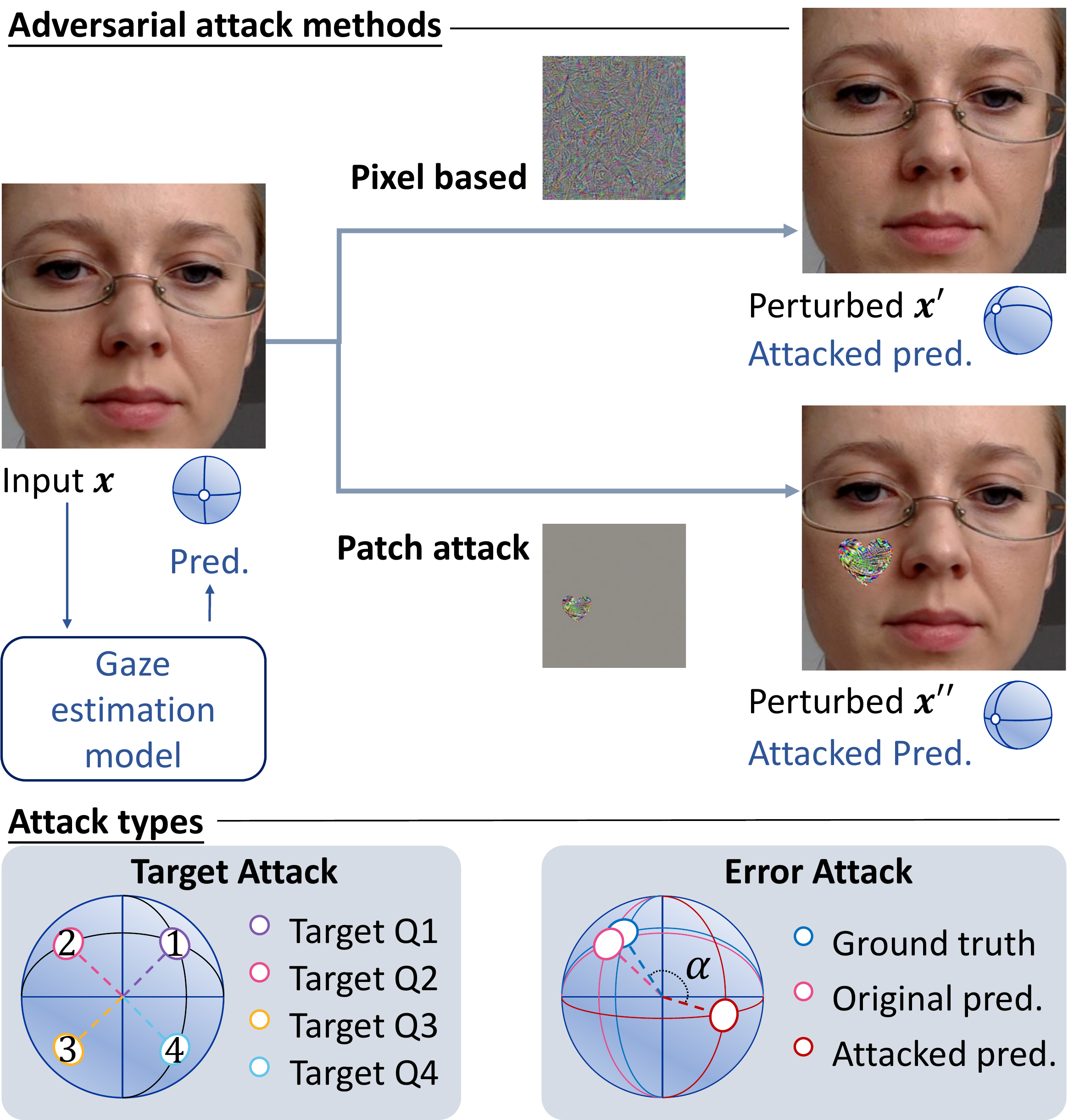}
   \end{center}
   \vspace{-4mm}
      \caption{The vulnerability of appearance-based gaze estimation.}
      \label{fig:vulnerability}
      \vspace{-6mm}
\end{figure}

Eye gaze is one of the important channel for communication. It indicates the region of interest during eye typing \cite{kurauchi2016eyeswipe, mott2017improving}, authentication \cite{khamis2016gazetouchpass, liu2015exploiting}, saliency prediction \cite{xu2016spatio}, object detection \cite{d2016gazed} and so on. Estimating gaze direction from eye appearance is challenging due to the diversity of human eyes. Conventional model-based gaze estimation approach seeks to recover the gaze direction from 3D eye model, but it usually requires dedicated devices such as infrared cameras. In contrast, the appearance-based gaze estimation approach only uses inexpensive web cameras, and directly learns the mapping function from eye/face images to gaze directions~\cite{lu2014adaptive,lu2015gaze}.

With the development of deep learning, the appearance-based gaze estimation accuracy has been significantly improved. A number of deep learning-based gaze estimation methods have been proposed. Sugano \etal  introduced learning-by-synthesis approach \cite{sugano2014learning} to appearance-based gaze estimation. Zhang \etal introduced the appearance-based gaze estimation method \cite{zhang15_cvpr} based on LeNet \cite{lecun1998gradient} using a grayscale single eye image as input. The backbone network of \cite{zhang15_cvpr} was changed to a 16-layer VGGNet \cite{simonyan2014very} afterwards, which was described in Zhang \etal \cite{DBLP:journals/corr/abs-1711-09017}. Zhang \etal used the full face image from the MPIIFaceGaze dataset to estimate gaze angles \cite{zhang2017s}. Cheng \etal explored the asymmetric of two eyes and improved the gaze estimation accuracy \cite{cheng2018appearance, cheng2020gaze}. Fischer \etal introduced a new dataset which makes ground truth annotation more accurately with eyetracking glasses, uses semantic image inpainting to remove glasses and presents a new real-time algorithm for gaze estimation \cite{FischerECCV2018}. Park \etal used very few examples to calibrate the model in order to learn the person-specific gazes \cite{park2019few}. Cheng \etal proposed a coarse-to-fine strategy \cite{cheng2020coarse} where the full face is used to obtain coarse-grained gaze direction and the combination of full face and two eyes are used for fine-grained gaze direction. Bao \etal recently proposed an effective method \cite{bao2020adaptive} that works well in mobile tablets.

However, the deep learning-based classification task often appears vulnerability \cite{szegedy2013intriguing, goodfellow2014explaining}, \ie, adding invisible noise on to the raw image leads to dramatic changes of network output. This problem can be summarized as following: given an input image $x$ that belongs to the target class $t$, we can easily generate the adversarial example $\widetilde{x}$ similar to $x$ to be classified as $t$. Szegedy \etal first discovered the vulnerability of deep learning models \cite{szegedy2013intriguing}. Goodfellow \etal used one-step fast gradient sign method \cite{goodfellow2014explaining} to attack the classification models. Kurakin \etal introduced Basic Iterative Method (BIM) \cite{kurakin2016physical}, which is an iterative method for attacking. Madry \etal used project gradient descent in the attack task \cite{madry2017towards}. Carlini \etal introduced attacks that are tailored to $L_2$, $L_\infty$ and $L_0$ distance metrics \cite{carlini2017towards}. Brown \etal  place a patch on images to make the images misclassified \cite{brown2017adversarial}, which is different from the attack methods described above.

In this paper, we investigate whether the vulnerability also exist in appearance-based gaze estimation task. By adding adversarial perturbation to the original inputs, we study if it is possible to change the predicted gaze direction, or even output a specific gaze direction. Fig. \ref{fig:vulnerability} shows the pipeline of this paper.

The main contributions of this paper are:
\begin{itemize}
   \item To our knowledge, we are the first to show the existence of the vulnerability in the appearance-based gaze-estimation tasks. We use the regression approach instead of classification to investigate the properties of vulnerability.
   \item We conduct a systematic study of vulnerability in terms of three aspects: the pixel-based adversarial attack, the patch-based adversarial attack and the defense strategy. We characterize the vulnerability from multiple aspects: hyper-parameters, face regions, model attention, smoothness, \etc.
   \item We report the performance of typical gaze estimation networks being attacked. We also evaluate the defense performance of these networks. Based on these results, we emphasize the necessity of defense in future appearance-based gaze estimation studies.
\end{itemize}

\section{Study overview}

\subsection{Study architecture}

This paper focuses on the vulnerability of appearance-based gaze estimation tasks. We characterize the vulnerability from two aspects, the attack side and the defense side. On the attack side, we investigate using pixel-based noise (Sec. \ref{section:pixel-attack}) and patch-based noise (Sec. \ref{section:patch-attack}) to attack the existing gaze estimation networks. We show the existence of the vulnerability and investigate the mechanism of vulnerability. We evaluate the performance of the attack under different parameter settings (Sec. \ref{section:pixel-performance-analysis}), and attack the different parts of the face (Sec. \ref{section:face-regions-effects}) to understand the roles of each face region during gaze estimation. We study the smoothness effect (Sec. \ref{section:smoothness}) to make the perturbations more invisible. To make the attack physically applicable, we investigate the patch-based adversarial attack and evaluated the performance of it (Sec. \ref{section:patch-performance-analysis}). We also explore how to defend against attacks, which is in Sec. \ref{section:study-on-defense}.

\subsection{Gaze estimation network}

To evaluate the vulnerability of appearance-based gaze estimation tasks, we select four typical gaze estimation networks, Full-Face \cite{zhang2017s}, Gaze-Net \cite{DBLP:journals/corr/abs-1711-09017}, CA-Net \cite{cheng2020coarse} and RT-GENE \cite{FischerECCV2018}. We introduce the details of these networks below. (The training data is randomly shuffled for each epoch.)

\textbf{Full-Face \cite{zhang2017s}.} This method is based on AlexNet \cite{krizhevsky2012imagenet} and uses the full face images as input. It is one of the most widely used method in appearance-based gaze estimation area. We used the MPIIFaceGaze dataset described in \cite{zhang2017s}. The batch size is $32$, the base learning rate is $0.01$ and weight\_decay is $0.0001$. We used SGD optimizer with momentum=0.9 and nesterov=True and multi-step scheduler with $\gamma=0.1$ and milestones=$[10,13]$(on epochs). The model is for $15$ epochs with $L_1$ loss.

\textbf{Gaze-Net \cite{DBLP:journals/corr/abs-1711-09017}.} This method is based on the pretrained VGG-16 \cite{simonyan2014very} and used grayscale single eye images as input. We use this method as it is the first CNN-based method for gaze estimation and often used as the baseline. We used MPIIGaze dataset described in \cite{DBLP:journals/corr/abs-1711-09017}. The batch size is $256$ and the base learning rate is $0.00001$. We used Adam optimizer with $\beta_1=0.9$ and $\beta_2=0.95$ and the StepLR scheduler with $step\_size=5000$ and $\gamma=0.1$. The model is trained for $20$ epochs with MSE loss. We used MSRA initialization \cite{he2015delving} to linear layers.

\textbf{CA-Net \cite{cheng2020coarse}.} CA-Net uses two eyes and the full face as input. Observed from the latter study, CA-Net shows superior performance against attack among the methods we used. This method is chosen as it achieves the state-of-the-art performance on MPIIFaceGaze \cite{zhang2017s} and EYEDIAP \cite{FunesMoraETRA2014} dataset (with Person \#12-\#13 removed as described in \cite{cheng2020coarse}). The batch size is $32$, the base learning rate to $0.001$ and $weight\_decay=0.1$. We used SGD optimizer. The model is trained for $100$ epochs. We used MSRA initialization \cite{he2015delving} to Conv2d layers.

\textbf{RT-GENE \cite{FischerECCV2018}.} This method has more accurate ground truth obtained by eye tracking glasses. We use ResNet-18 \cite{he2016deep} backbone model due to the low computation cost. This method used two gray-scale eye images as input. We have trained two models with (described as RT-GENE) and without (described as RT-GENE (Augmented)) data augmentation, respectively. The batch size is $128$ and base learning rate is $0.000325$. We trained the model with MSE loss. The model is trained for 5 epochs if data augmentation is disabled or 64.

\subsection{Attack task definition}

Our goal is to confuse the gaze estimation model and output the certain target gaze direction. Given an appearance-based gaze estimator $\mathcal{G}$ and the 3D target gaze direction $\mathbf{t} \in \mathbb{R}^3$, we generate the adversarial example $\widetilde{\mathbf{x}}$ by minimizing the angular error $\mathcal{L}$ between the estimated gaze direction $\mathcal{G}(\widetilde{\mathbf{x}}, \dots)$ and the target direction $\mathbf{t}$, as shown below:
\begin{equation}
   \begin{split}
      \text{min } \;& \mathcal{L}(\mathcal{G}(\widetilde{\mathbf{x}}, \dots), \mathbf{t})\\
      \text{s.t. } \;&\| \widetilde{\mathbf{x}}-\mathbf{x} \|_\infty \le \epsilon, \widetilde{\mathbf{x}} \in [0,255]^n, \label{eq:optimization}
   \end{split}
\end{equation}
where the angular error $\mathcal{L}$ is defined as:
\begin{equation}
   \begin{split}
      \mathcal{L}(\mathcal{G}(\widetilde{\mathbf{x}}, \dots), \mathbf{t}) = \cos^{-1}{\left( \frac{\mathbf{t} \cdot \mathcal{G}(\widetilde{\mathbf{x}}, \dots)}{ \| \mathbf{t}\| \cdot \| \mathcal{G}(\widetilde{\mathbf{x}}, \dots) \|} \right)}. \label{eq:mean_angle_loss}
   \end{split}
\end{equation}

Note that we have used $\mathcal{G}(\widetilde{\mathbf{x}}, \dots)$ instead of $\mathcal{G}(\widetilde{\mathbf{x}})$ since we may have multiple inputs fed into the gaze estimator. $\| \widetilde{\mathbf{x}}-\mathbf{x} \|_\infty \le \epsilon$ constrains the $L_\infty$ distance between $\widetilde{\mathbf{x}}$ and $\mathbf{x}$ to make $\widetilde{\mathbf{x}}$ similar to $\mathbf{x}$.

In this paper, we use the angular error $\mathcal{L}$ between \textbf{the estimated gaze direction $\mathcal{G}$ and the target gaze direction $\mathbf{t}$} to measure the vulnerability of the gaze estimation networks. The unit of angular error is degree. The smaller $\mathcal{L}$ is, the more vulnerable the method is, since the network outputs can easily changed. Meanwhile, we also measure angular error $\mathcal{L}'$ between \textbf{the estimated gaze direction $\mathcal{G}$ and the ground truth gaze direction}. Large $\mathcal{L}'$ means that the estimated gaze direction after attack is far from the ground truth, which also indicates the vulnerability.


Without loss of generality, we denote the 2D target gaze directions as ($pitch$, $yaw$) and select four target gaze directions in the target attack task. As shown in the right part of Fig. \ref{fig:vulnerability}, we use $T_{Q1}$, $T_{Q2}$, $T_{Q3}$, and $T_{Q4}$ to represent four target directions $(\frac{\pi}{4},\frac{\pi}{4}), (-\frac{\pi}{4},\frac{\pi}{4}), (-\frac{\pi}{4},-\frac{\pi}{4})$ and $(\frac{\pi}{4},-\frac{\pi}{4})$. Each dataset is divided into several folds according to the person. We follow the leave-one-person-out strategy to train and test the model. For example, we train on the Person \#1-\#14 of MPIIGaze, then test and attack on Person \#0. If no special instruction, we only use one fold of the dataset (Person \#0 for MPIIFaceGaze and MPIIGaze, Person \#1 for EYEDIAP, Fold \#0 for RT-GENE) for each method to attack, and the other folds to train the model. The final result is the mean angle loss across these four target directions and all the inputs inside the corresponding fold.

In practice, a batch of input images $\mathbf{x}$ instead of one input image will be used for the optimization task. If there are more than one type of output, \eg, the coarse output and the fine output, we use the sum of angular error $\mathcal{L}$ of each type of output. If there are more than one type of input, \eg, the eyes and the face, each type of input will be attacked independently.

\section{Study on pixel-based adversarial attack}
\label{section:pixel-attack}

\subsection{Method}
\label{section:pixel-method}
The pixel-based adversarial attack is the modified from the Basic Iterative Method (BIM) \cite{kurakin2016adversarial}, which is described as follows:
\begin{equation}
   \begin{split}
      \mathbf{x}_{k} = Clip_{\mathbf{x}, \epsilon}\Bigl\{ \mathbf{x}_{k-1} - \alpha \; \text{sign} \bigl( \nabla_\mathbf{x} \mathcal{L}(\mathcal{G}(\mathbf{x}_{k-1}, \dots), \mathbf{t})  \bigr) \Bigr\}, \label{eq:bim}
   \end{split}
\end{equation}
where $k\in [1,N]$ is the iteration index, $Clip_{\mathbf{x}, \epsilon}(\mathbf{A}_{i,j})$ means clipping $\mathbf{A}_{i,j}$ to the range $[\mathbf{x}_{i, j}-\epsilon, \mathbf{x}_{i, j}+\epsilon]$ that meets the $\| \widetilde{\mathbf{x}}-\mathbf{x} \|_\infty \le \epsilon$ constriant. The final adversarial example $\widetilde{\mathbf{x}}$ is the $\mathbf{x}_{k}$ that minimizes $\mathcal{L}(\mathcal{G}(\mathbf{x}_{k}, \dots), \mathbf{t})$.

There are three hyper-parameters, $\epsilon$, $\alpha$ and $N$, where $\epsilon$ represents the intensity of the $\| \widetilde{\mathbf{x}}-\mathbf{x} \|_\infty$ constraint, $\alpha$ is the stride of one iteration, and $N$ is the number of iterations. The original paper \cite{kurakin2016adversarial} sets $\alpha=1$ and $N=\min(\epsilon+4,1.25\epsilon)$. Here we modified $N$ to $\lfloor \max{\left( {\epsilon}/{\alpha}+4, 2 \cdot {\epsilon}/{\alpha} \right) \rceil}$ to guarantee fairness for different $\epsilon$ and $\alpha$.

\subsection{Characterization}

\subsubsection{Performance analysis}
\label{section:pixel-performance-analysis}

First, we evaluate the effects of model parameters $\epsilon$ and $\alpha$. The testing range of $\epsilon$ is $\left\{  1,2,4,8,16,32,64 \right\}$, the testing range of $\alpha$ is $\left\{ 0.125,0.25,0.5,1,2,4 \right\}$. In theory, the maximum possible value of $\epsilon$ is 255. Here we choose the maximum value of $\epsilon$ as 64 since it is large enough to generate the typical attack results. In the original paper \cite{kurakin2016adversarial}, the authors set $\alpha$ to be 1. Here we test a wider range of $\alpha$ from 0.125 to 4 to see the influence of $\alpha$ on attack task. The results of mean angular error under different combinations of $\epsilon$ and $\alpha$ can be found in Table \ref{table:hyper_parameters}. The ``Heatmaps'' in the right column of the table visualize the changing trend.

We observe that the angular error increases as the stride size $\alpha$ increases. This may be due to that smaller stride leads the model to be attacked more granularly while the larger stride makes the loss more difficult to converge. For Full-Face, Gaze-Net and RT-GENE, the mean angular error significantly decreases as $\epsilon$ increases. In other words, the network becomes easier to be attacked. It can be inferred that these models have strong vulnerability. However, the mean angular error of CA-Net decays very slowly when testing on MPIIFaceGaze and EYEDIAP, which indicates that this model is robust. From the heatmaps in Table \ref{table:hyper_parameters}, we also observe that when $\epsilon$ increases, the mean angular errors of different methods decay at different speeds. This indicates different vulnerability of different methods in terms of $\epsilon$: Full-Face $>$ RT-GENE $>$ RT-GENE (Augmented) $>$ Gaze-Net $>$ CA-Net. This conclusion suggest that the CA-Net is most robust against the attacks.


\begin{table*}[!hbt]
\caption{Evaluation results of hyper-parameter effects}
\label{table:hyper_parameters}
\vspace{-3mm}
\centering
\small{{
   \begin{tabular}{c|c|ccccccc|c}
      \toprule
      \multirow{2}[2]{*}{\shortstack{\textbf{Dataset}\\\textbf{+ Method}}} & \multirow{2}[2]{*}{$\alpha$} & \multicolumn{7}{c}{
         \textbf{Mean angular error when} $\epsilon=$
      } & \multirow{2}[2]{*}{\textbf{Heatmap of angular error}} \\
      & & 1 & 2 & 4 & 8 & 16 & 32 & 64 & \\
      \midrule
      \multirow{6}[2]{*}{\shortstack{MPIIFaceGaze \cite{zhang2017s}\\+ Full-Face \cite{zhang2017s}}} 
          & 4 & 44.97 & 38.52 & 19.16 & 10.05 & 8.933 & 8.932 & 8.868 & \multirow{6}[2]{*}{
            \begin{minipage}[b]{0.35\columnwidth}
               \centering
               \raisebox{-.5\height}{\includegraphics[width=1.0\linewidth]{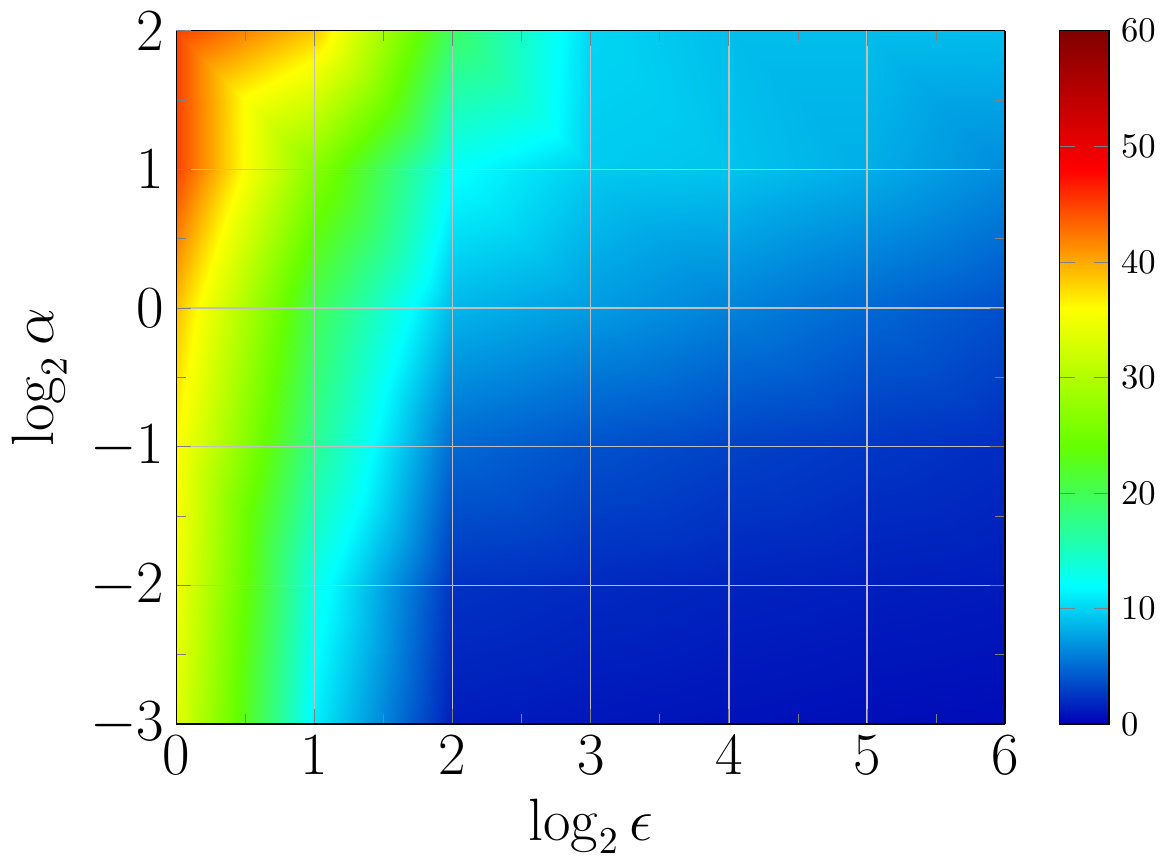}}
            \end{minipage}
         }  \\
         & 2 & 44.97 & 26.60 & 12.29 & 9.533 & 9.221 & 8.204 & 6.624 & \\
         & 1 & 39.01 & 20.08 & 8.603 & 7.337 & 6.158 & 4.918 & 3.759 & \\
         & 0.5 & 36.11 & 17.13 & 4.911 & 3.812 & 3.110 & 2.569 & 2.098 & \\
         & 0.25 & 35.14 & 13.05 & 2.367 & 1.956 & 1.648 & 1.421 & 1.242 & \\
         & 0.125 & 33.79 & 11.24 & 1.295 & 1.048 & 0.852 & 0.674 & 0.551 & \\
        \midrule
        \multirow{6}[2]{*}{\shortstack{RT-GENE \cite{FischerECCV2018}}} 
            & 4      & 51.37 & 49.04 & 38.22 & 28.29 & 20.34 & 12.39 & 8.001 & \multirow{6}[2]{*}{
            \begin{minipage}[b]{0.35\columnwidth}
                \centering
                \raisebox{-.5\height}{\includegraphics[width=1.0\linewidth]{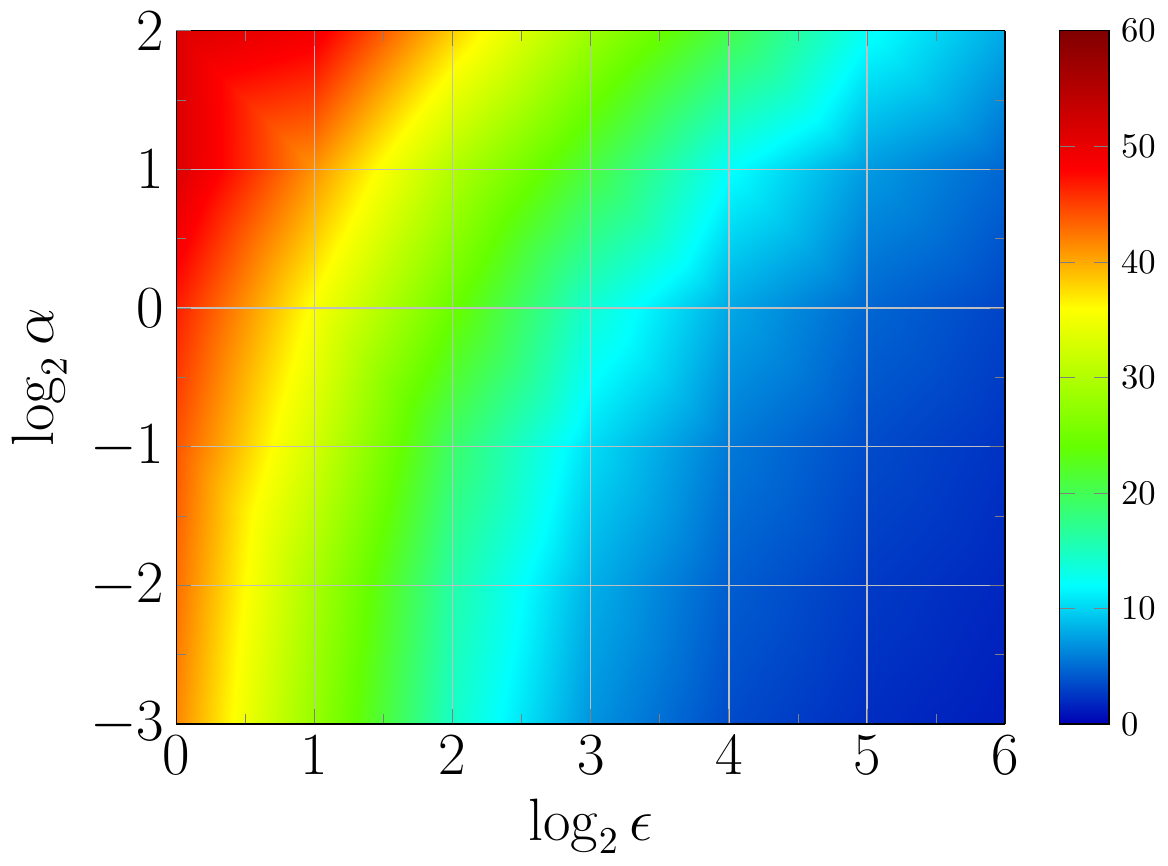}}
            \end{minipage}
        }  \\
        & 2      & 51.37 & 40.97 & 29.94 & 20.89 & 12.23 & 7.282 & 4.812 & \\
        & 1      & 46.67 & 35.50 & 24.44 & 14.05 & 7.907 & 4.812 & 3.207 & \\
        & 0.5    & 44.06 & 32.42 & 19.01 & 10.36 & 5.716 & 3.555 & 2.371 & \\
        & 0.25   & 42.84 & 29.22 & 16.05 & 8.261 & 4.401 & 2.744 & 1.810 & \\
        & 0.125  & 41.66 & 27.74 & 14.56 & 6.977 & 3.522 & 2.100 & 1.326 & \\
        \midrule
        \multirow{6}[2]{*}{\shortstack{RT-GENE \cite{FischerECCV2018}\\(Augmented)}} 
            & 4      & 49.73 & 48.88 & 41.59 & 34.33 & 26.42 & 14.71 & 8.527 & \multirow{6}[2]{*}{
            \begin{minipage}[b]{0.35\columnwidth}
                \centering
                \raisebox{-.5\height}{\includegraphics[width=1.0\linewidth]{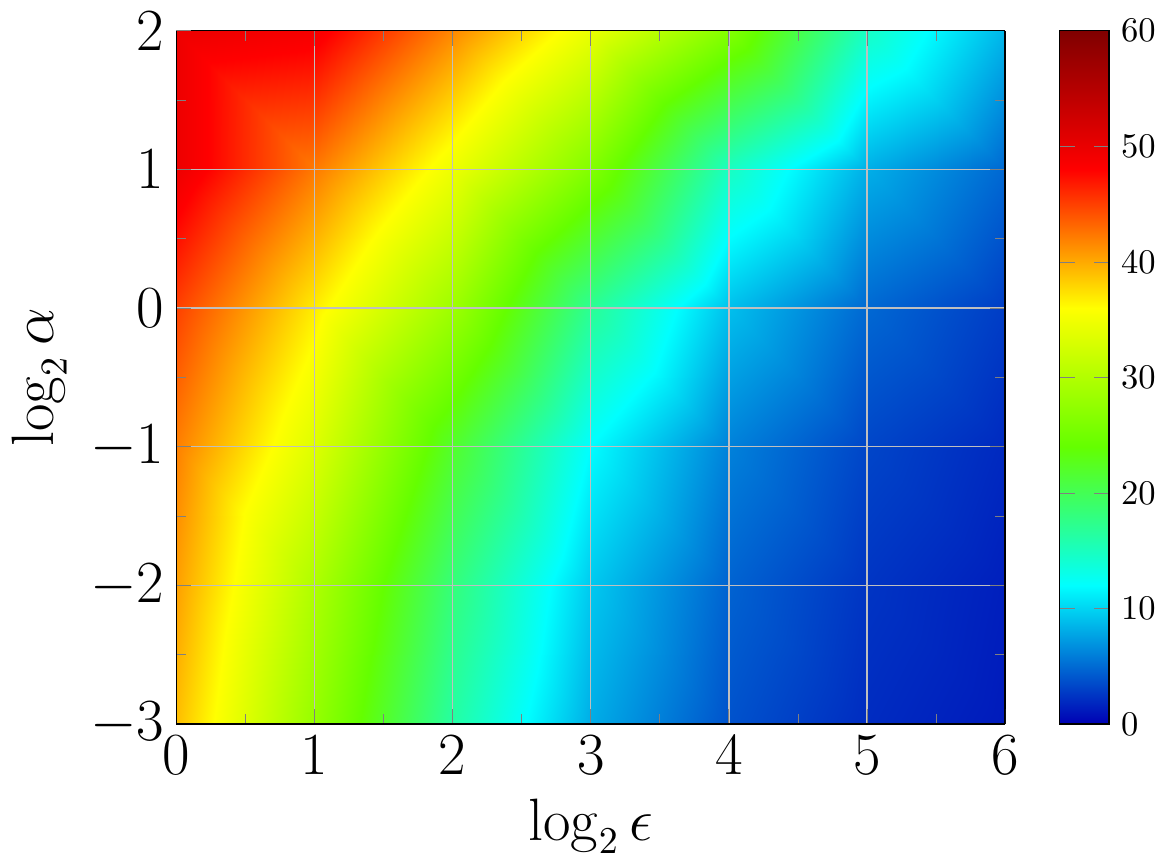}}
            \end{minipage}
        }  \\
        & 2      & 49.73 & 42.26 & 34.66 & 26.33 & 14.89 & 8.218 & 4.870 & \\
        & 1      & 45.02 & 36.82 & 28.43 & 17.00 & 9.006 & 4.905 & 2.846 & \\
        & 0.5    & 42.05 & 33.35 & 21.88 & 11.92 & 6.081 & 3.265 & 1.919 & \\
        & 0.25   & 40.56 & 29.67 & 18.23 & 9.433 & 4.644 & 2.479 & 1.444 & \\
        & 0.125  & 39.11 & 27.74 & 16.29 & 8.101 & 3.861 & 2.047 & 1.235 & \\
      \midrule
      \multirow{6}[2]{*}{\shortstack{MPIIGaze \cite{DBLP:journals/corr/abs-1711-09017}\\+ Gaze-Net \cite{DBLP:journals/corr/abs-1711-09017}}} 
          & 4 & 57.62 & 55.55 & 51.38 & 44.08 & 34.24 & 20.37 & 7.570 & \multirow{6}[2]{*}{
            \begin{minipage}[b]{0.35\columnwidth}
               \centering
               \raisebox{-.5\height}{\includegraphics[width=1.0\linewidth]{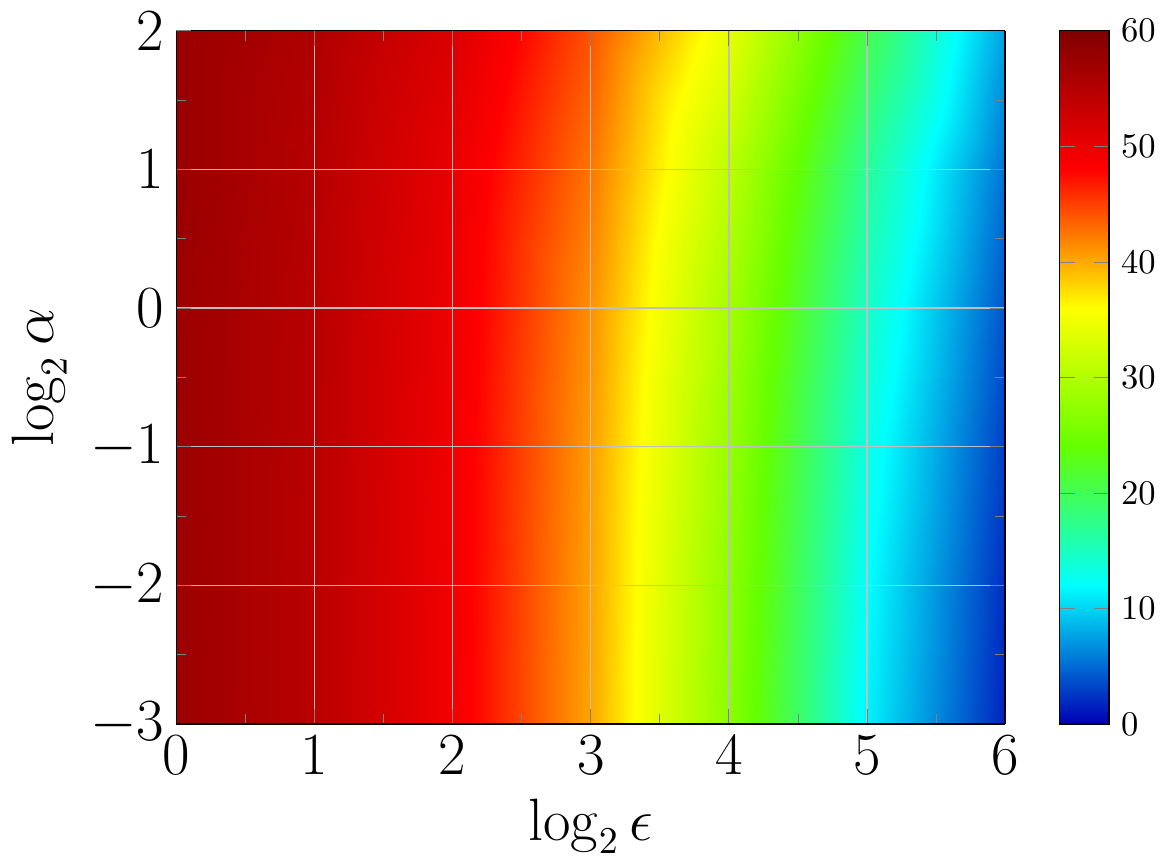}}
            \end{minipage}
         }  \\
         & 2 & 57.62 & 54.94 & 50.07 & 42.45 & 30.85 & 16.83 & 5.354 & \\
         & 1 & 57.47 & 54.65 & 49.67 & 41.21 & 29.01 & 14.75 & 3.982 & \\
         & 0.5 & 57.43 & 54.61 & 49.38 & 40.61 & 27.94 & 13.24 & 2.918 & \\
         & 0.25 & 57.43 & 54.55 & 49.25 & 40.31 & 27.29 & 12.13 & 2.117 & \\
         & 0.125 & 57.42 & 54.53 & 49.21 & 40.18 & 26.93 & 11.39 & 1.617 & \\
      \midrule
      \multirow{6}[2]{*}{\shortstack{EYEDIAP \cite{FunesMoraETRA2014}\\+ CA-Net \cite{cheng2020coarse}}} 
          & 4      & 52.28  & 50.16  & 45.77  & 43.04  & 41.40  & 39.91  & 39.19 & \multirow{6}[2]{*}{
            \begin{minipage}[b]{0.35\columnwidth}
               \centering
               \raisebox{-.5\height}{\includegraphics[width=1.0\linewidth]{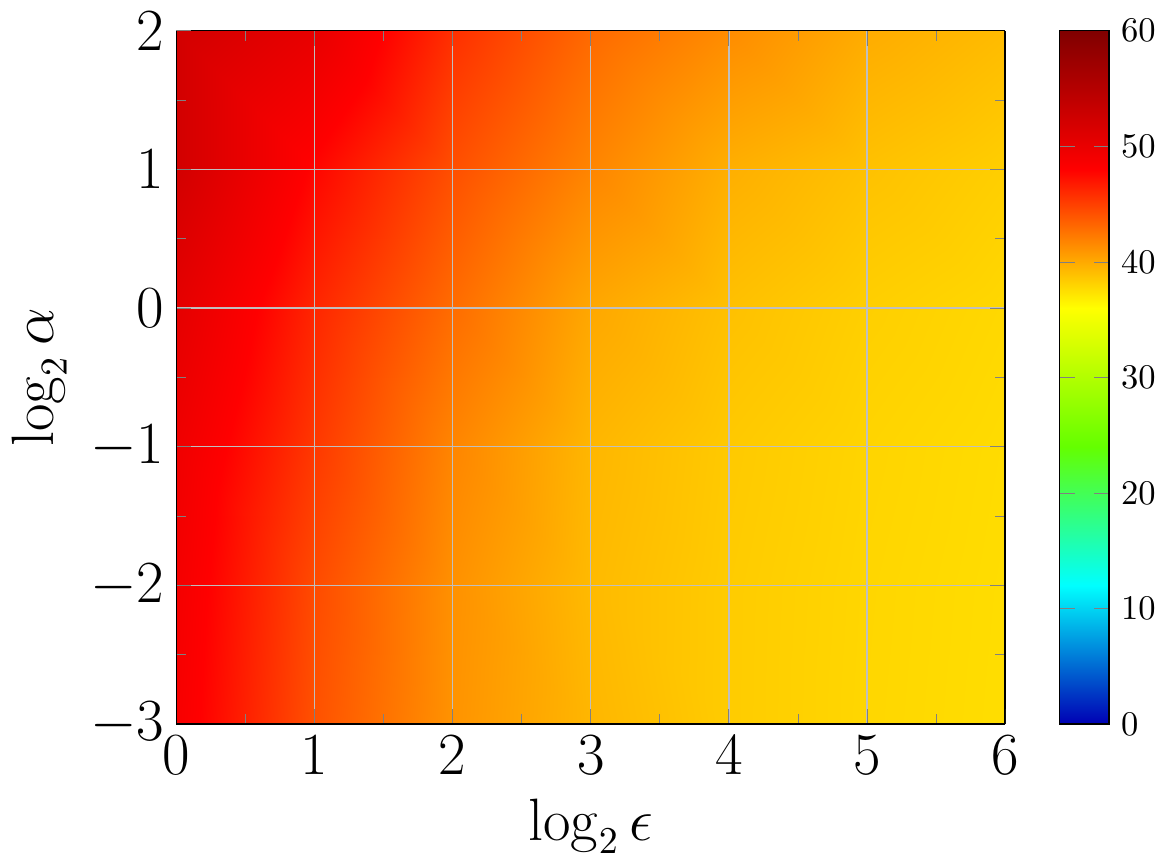}}
            \end{minipage}
         }  \\
         & 2      & 52.28  & 47.75  & 44.29  & 41.73  & 39.77  & 38.90  & 38.29 & \\
         & 1      & 50.55  & 46.22  & 42.91  & 40.12  & 38.94  & 38.26  & 37.84 & \\
         & 0.5    & 49.52  & 45.35  & 41.72  & 39.42  & 38.53  & 38.01  & 37.66 & \\
         & 0.25   & 49.06  & 44.48  & 41.18  & 39.25  & 38.43  & 37.93  & 37.60 & \\
         & 0.125  & 48.78  & 44.17  & 41.07  & 39.42  & 38.42  & 37.90  & 37.57 & \\
      \midrule
      \multirow{6}[2]{*}{\shortstack{MPIIFaceGaze \cite{zhang2017s}\\+ CA-Net \cite{cheng2020coarse}}} 
          & 4      & 53.81  & 51.64  & 45.74  & 42.03  & 40.09  & 38.72  & 38.08 & \multirow{6}[2]{*}{
            \begin{minipage}[b]{0.35\columnwidth}
               \centering
               \raisebox{-.5\height}{\includegraphics[width=1.0\linewidth]{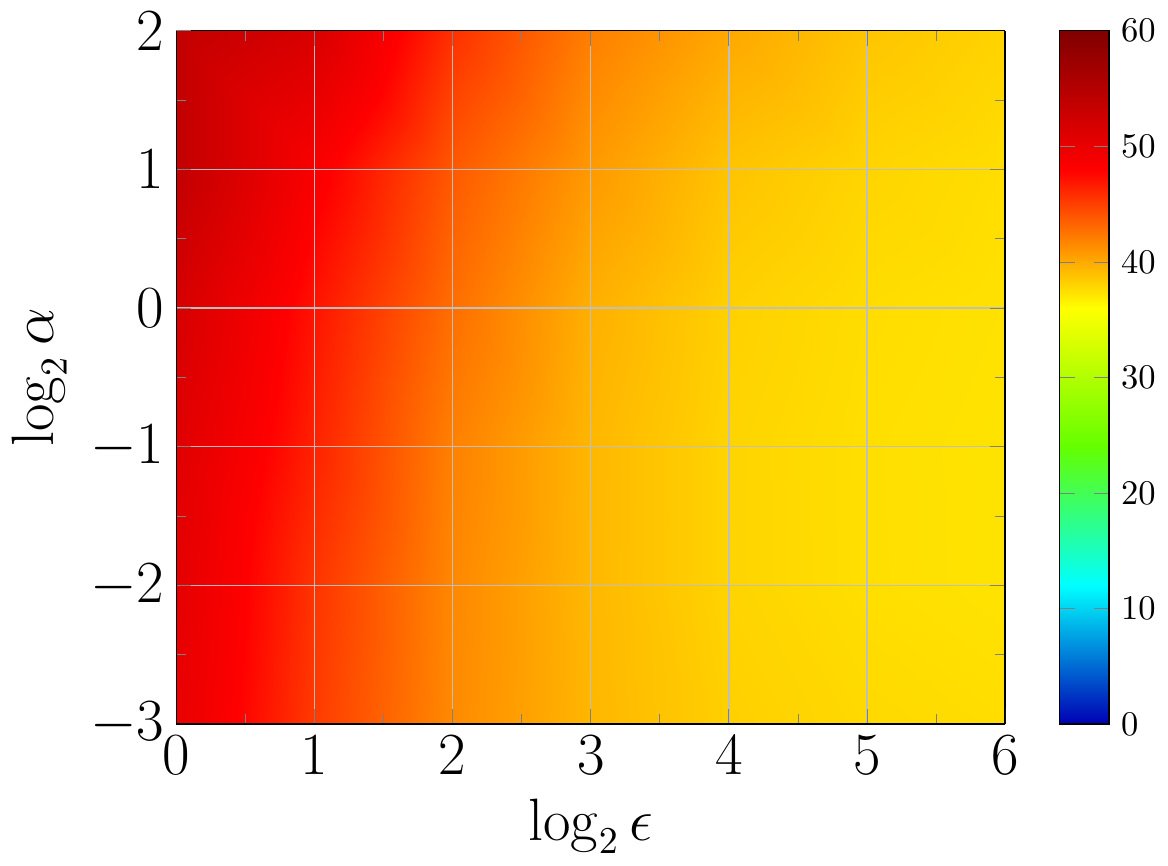}}
            \end{minipage}
         }  \\
         & 2      & 53.81  & 48.56  & 43.82  & 40.76  & 38.79  & 38.00  & 37.57 & \\
         & 1      & 51.88  & 47.07  & 42.80  & 39.72  & 38.19  & 37.62  & 37.38 & \\
         & 0.5    & 50.99  & 46.39  & 41.93  & 39.38  & 38.01  & 37.51  & 37.29 & \\
         & 0.25   & 50.54  & 45.53  & 41.60  & 39.34  & 38.02  & 37.52  & 37.30 & \\
         & 0.125  & 50.14  & 45.23  & 41.49  & 39.40  & 38.23  & 37.80  & 37.59 & \\
      \bottomrule
   \end{tabular}}}
\vspace{-4mm}
\end{table*}

\begin{figure}[t]
   \subfigure[$\epsilon=1, \alpha=0.125$]{
      \begin{minipage}[b]{0.3\linewidth}
         \includegraphics[width=1.0\linewidth]{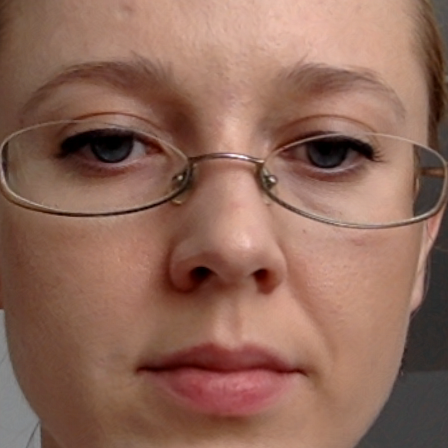}
      \end{minipage}
      \label{fig:epsilon-0.125-1}
   }
   \subfigure[$\epsilon=8, \alpha=0.125$]{
      \begin{minipage}[b]{0.3\linewidth}
         \includegraphics[width=1.0\linewidth]{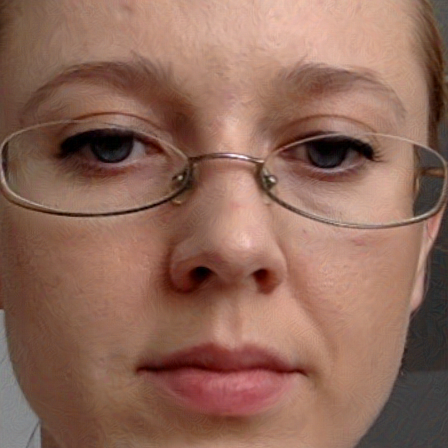}
      \end{minipage}
      \label{fig:epsilon-0.125-8}
   }
   \subfigure[$\epsilon=64, \alpha=0.125$]{
      \begin{minipage}[b]{0.3\linewidth}
         \includegraphics[width=1.0\linewidth]{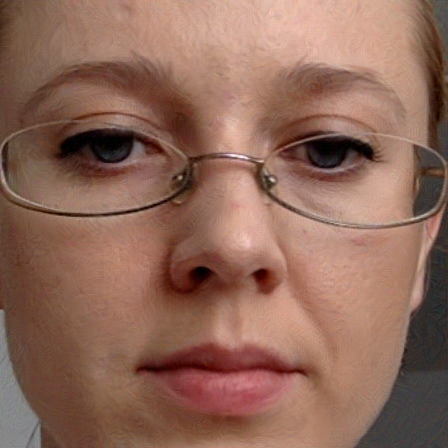}
      \end{minipage}
      \label{fig:epsilon-0.125-64}
   }
   \subfigure[$\epsilon=1, \alpha=1$]{
      \begin{minipage}[b]{0.3\linewidth}
         \includegraphics[width=1.0\linewidth]{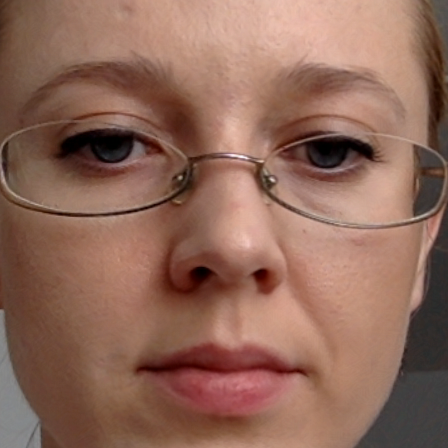}
      \end{minipage}
      \label{fig:epsilon-1-1}
   }
   \subfigure[$\epsilon=8, \alpha=1$]{
      \begin{minipage}[b]{0.3\linewidth}
         \includegraphics[width=1.0\linewidth]{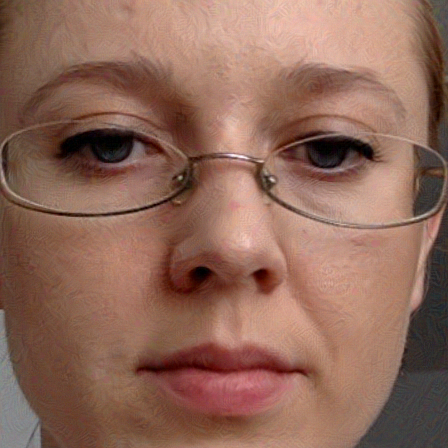}
      \end{minipage}
      \label{fig:epsilon-1-8}
   }
   \subfigure[$\epsilon=64, \alpha=1$]{
      \begin{minipage}[b]{0.3\linewidth}
         \includegraphics[width=1.0\linewidth]{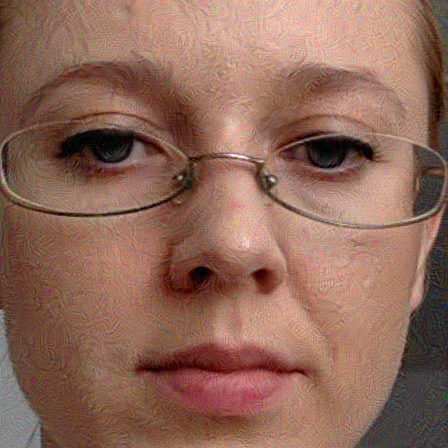}
      \end{minipage}
      \label{fig:epsilon-1-64}
   }
   \subfigure[$\epsilon=1, \alpha=4$]{
      \begin{minipage}[b]{0.3\linewidth}
         \includegraphics[width=1.0\linewidth]{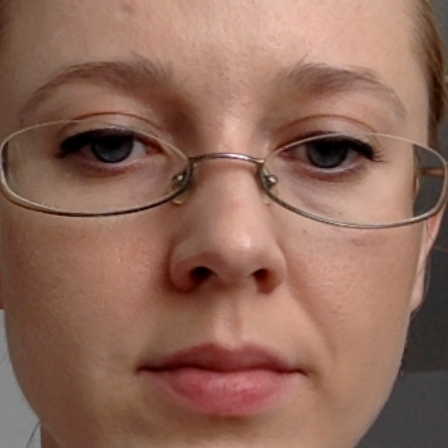}
      \end{minipage}
      \label{fig:epsilon-4-1}
   }
   \subfigure[$\epsilon=8, \alpha=4$]{
      \begin{minipage}[b]{0.3\linewidth}
         \includegraphics[width=1.0\linewidth]{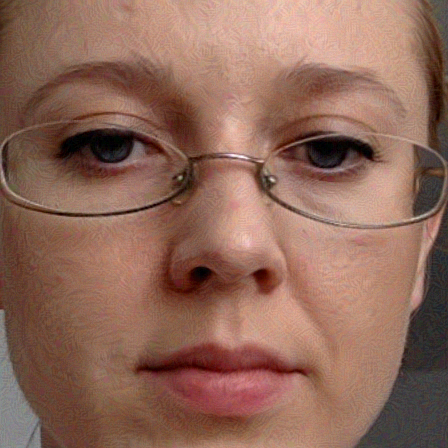}
      \end{minipage}
      \label{fig:epsilon-4-8}
   }
   \subfigure[$\epsilon=64, \alpha=4$]{
      \begin{minipage}[b]{0.3\linewidth}
         \includegraphics[width=1.0\linewidth]{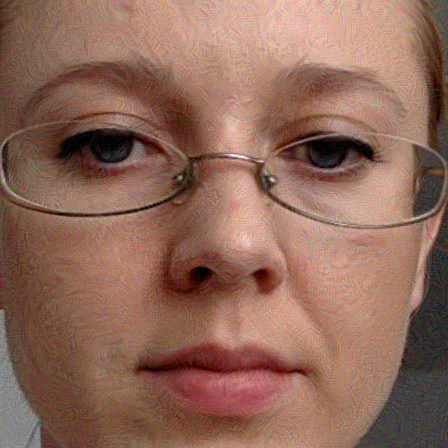}
      \end{minipage}
      \label{fig:epsilon-4-64}
   }
   \caption{Perturbed images under different ($\epsilon,\alpha$).}
\label{fig:epsilon}
\end{figure}

Figure \ref{fig:epsilon} visualizes the perturbed images under different $\epsilon$. Since $\epsilon$ represents the intensity of $L_\infty$ constriant, it can be inferred that greater $\epsilon$ may lead to farther distance between $\widetilde{\mathbf{x}}$ and $\mathbf{x}$, and the perturbations may become more obvious. For example, when $\alpha=1$ and $4$ (Fig. \ref{fig:epsilon-1-1}, \ref{fig:epsilon-1-8}, \ref{fig:epsilon-1-64}, \ref{fig:epsilon-4-1}, \ref{fig:epsilon-4-8}, \ref{fig:epsilon-4-64}), the perturbation become more obvious as $\epsilon$ getting larger. However, in some cases this conclusion is false. For example, when $\alpha=0.125$ (\ref{fig:epsilon-0.125-1}, \ref{fig:epsilon-0.125-8}, \ref{fig:epsilon-0.125-64}), the perturbation is not obvious even though $\epsilon$ is high enough.


We also investigate the wide vulnerability of models. While the experiments above are all limited to a single fold of dataset, we also conduct experiments attacking on each folds for every model. As $\epsilon=64$ and $\alpha=0.125$ get the most best results in Table \ref{table:hyper_parameters}, we choose this combination of $\epsilon$ and $\alpha$. For each model we compute the mean and standard deviation of angular error across all inputs and targets. The results are displayed in Table \ref{table:universality}. We observed that the angular error of Full-Face, Gaze-Net and RT-GENE are all close to $0$, indicating that the estimated gaze directions for these models are easily to be changed. This result shows the wide and strong vulnerability of these models.

\begin{table}[!hbt]
\caption{Evaluation results of the wide vulnerability. ``Mean'' is the mean angular error across inputs and targets in all folds, ``Std'' is the standard deviation of angular error across all inputs and targets in all folds.}
\label{table:universality}
\centering
\small{{
   \begin{tabular}{c|c}
      \toprule
      \textbf{Dataset+Method} & \textbf{Mean angular error} \\
      \midrule
      \shortstack{MPIIFaceGaze\cite{zhang2017s}+Full-Face\cite{zhang2017s}}
      & 0.60$\pm$0.48 \\
      \midrule
      \shortstack{MPIIGaze\cite{DBLP:journals/corr/abs-1711-09017}+Gaze-Net\cite{DBLP:journals/corr/abs-1711-09017}}
      & 0.64$\pm$1.76 \\
      \midrule
      \shortstack{EYEDIAP \cite{FunesMoraETRA2014}+ CA-Net\cite{cheng2020coarse}}
      & 37.40$\pm$7.14 \\
      \midrule
      \shortstack{MPIIFaceGaze \cite{zhang2017s}+ CA-Net\cite{cheng2020coarse}}
      & 36.52$\pm$10.12 \\
      \midrule
      \shortstack{RT-GENE \cite{FischerECCV2018}}
      & 1.77$\pm$3.95 \\
      \midrule
      \shortstack{RT-GENE \cite{FischerECCV2018}(Augmented)}
      & 3.21$\pm$5.78 \\
      \bottomrule
   \end{tabular}}}
\end{table}

However, the angular error of CA-Net is still very high even though $\epsilon$ becomes very large. We found that the standard deviation of CA-Net is larger than that of other models, so we show the results of CA-Net within each target in Table \ref{table:std_target}. We also choose $\epsilon=64$ and $\alpha=0.125$. It can be found that although the standard deviation of CA-Net in \ref{table:universality} is very large, the difference within one target is relatively small, which indicates that towards a certain target, the estimated gaze directions after attack are centered at a value that is far from the target. This result clearly illustrates the robustness of CA-Net.


\begin{table}[!hbt]
\caption{Angular error within each target.}
\label{table:std_target}
\centering
\small{\resizebox{\linewidth}{!}{
   \begin{tabular}{c|cccc}
      \toprule
      \textbf{Dataset+Method} & Target Q1 & Target Q2 & Target Q3 & Target Q4 \\
      \midrule
      \shortstack{EYEDIAP \cite{FunesMoraETRA2014}+CA-Net\cite{cheng2020coarse}} 
      & 30.74$\pm$2.19 & 32.22$\pm$3.79 & 43.91$\pm$4.85 & 42.73$\pm$4.41 \\
      \midrule
      \shortstack{MPIIFaceGaze \cite{zhang2017s}+CA-Net\cite{cheng2020coarse}}
      & 46.75$\pm$3.88 & 45.71$\pm$3.47 & 27.32$\pm$1.54 & 26.31$\pm$1.34 \\
      \bottomrule
   \end{tabular}}}
\end{table}

\subsubsection{Face regions effects}
\label{section:face-regions-effects}

\begin{figure}[t]
   \begin{center}
      \begin{minipage}[b]{0.46\linewidth}
         \includegraphics[width=1.0\linewidth]{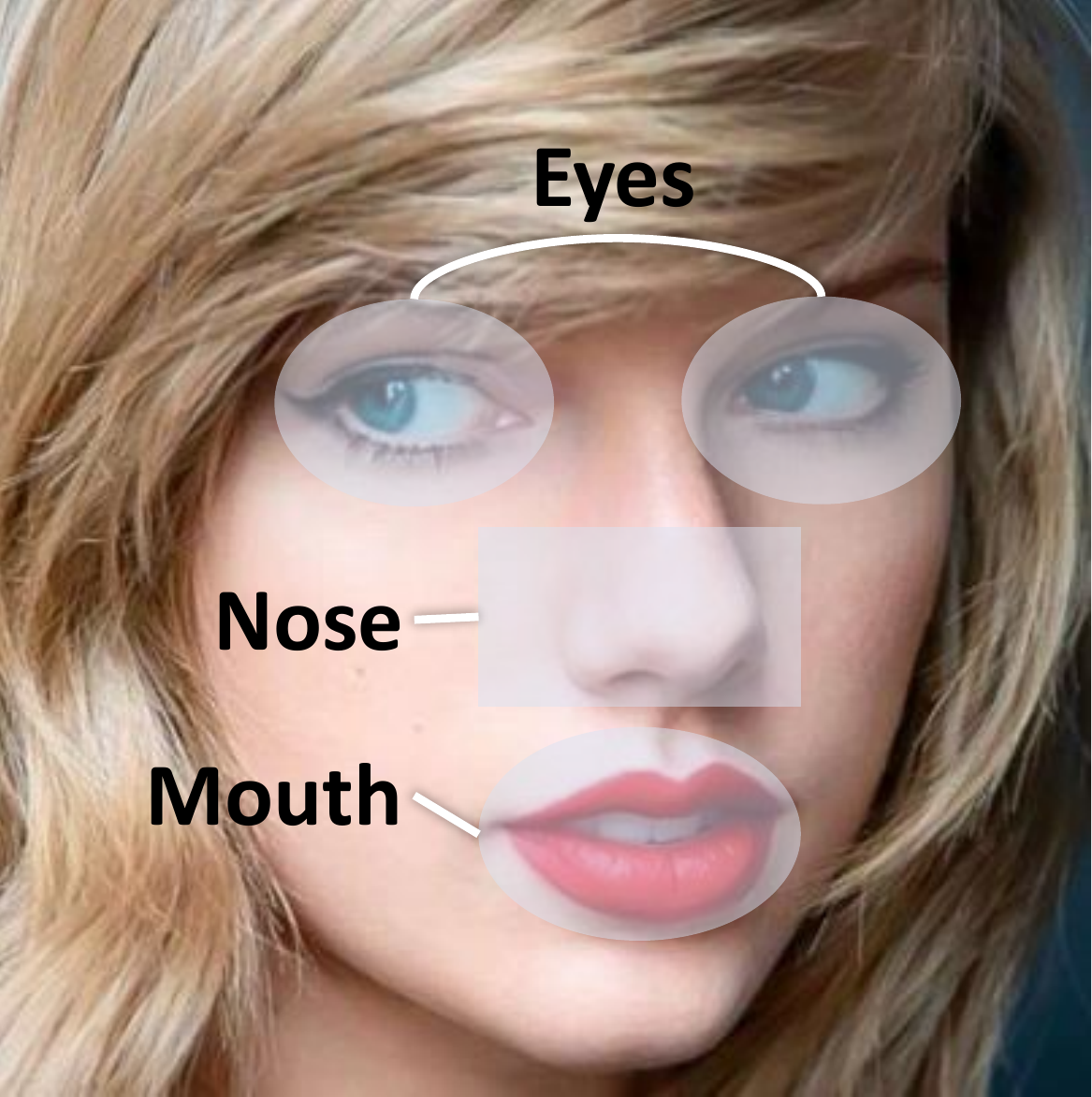}
      \end{minipage}
      \vspace{-1em}
   \end{center}
      \caption{The definition of different face parts.}
      \label{fig:face_parts}
      \vspace{-1em}
\end{figure}

\begin{figure}[t]
   \centering
   \subfigure[Eyes]{
      \begin{minipage}[b]{0.3\linewidth}
         \begin{overpic}[width=1\textwidth]{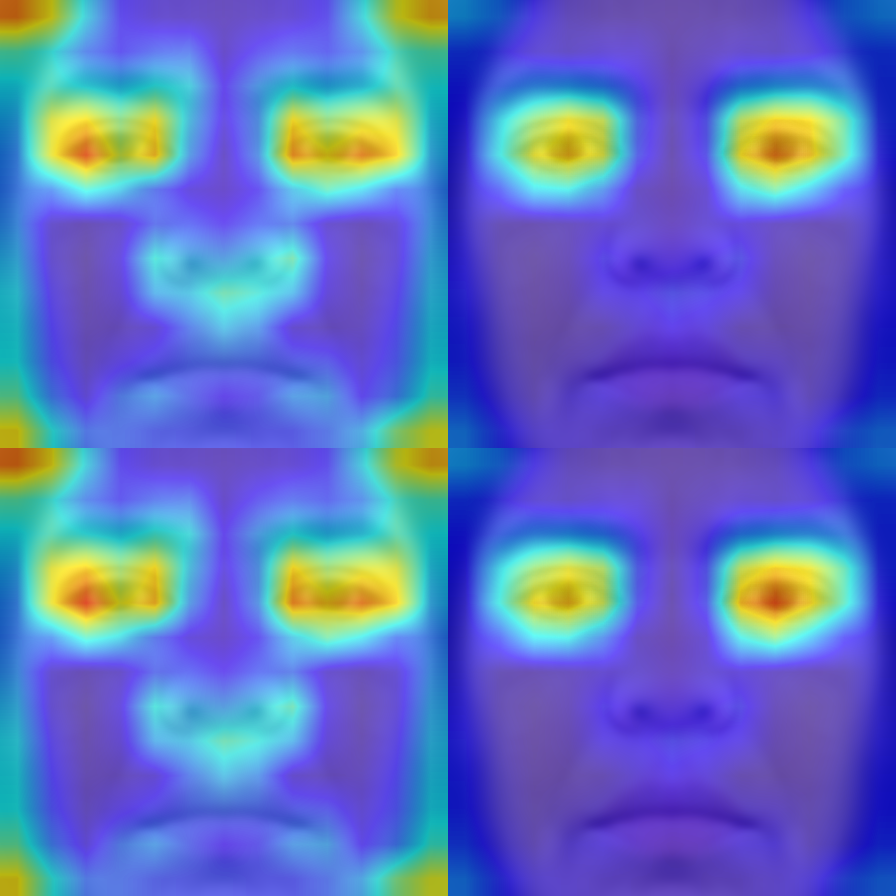}
            \put(3, 52){\tiny\color{white}{before, $pitch$}}
            \put(56, 52){\tiny\color{white}{after, $pitch$}}
            \put(6, 2){\tiny\color{white}{before, $yaw$}}
            \put(59, 2){\tiny\color{white}{after, $yaw$}}
         \end{overpic}
      \end{minipage}
      \label{fig:attention_eyes}
   }
   \subfigure[Nose]{
      \begin{minipage}[b]{0.3\linewidth}
         \begin{overpic}[width=1\textwidth]{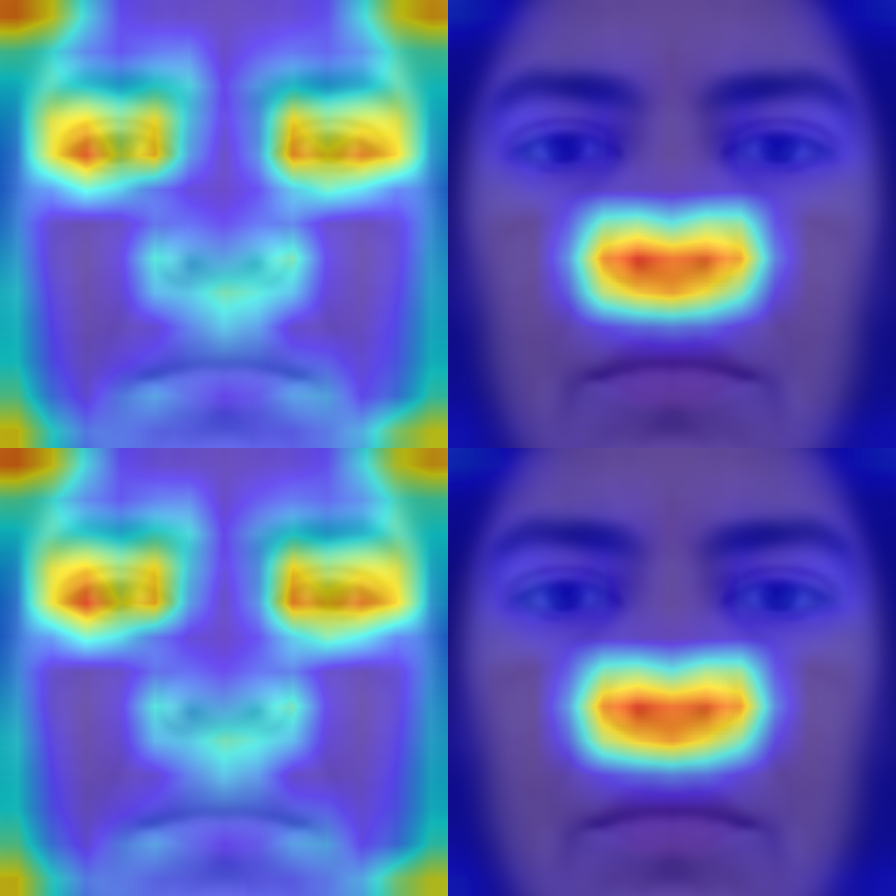}
            \put(3, 52){\tiny\color{white}{before, $pitch$}}
            \put(56, 52){\tiny\color{white}{after, $pitch$}}
            \put(6, 2){\tiny\color{white}{before, $yaw$}}
            \put(59, 2){\tiny\color{white}{after, $yaw$}}
         \end{overpic}
      \end{minipage}
      \label{fig:attention_nose}
   }
   \subfigure[Mouth]{
      \begin{minipage}[b]{0.3\linewidth}
         \begin{overpic}[width=1\textwidth]{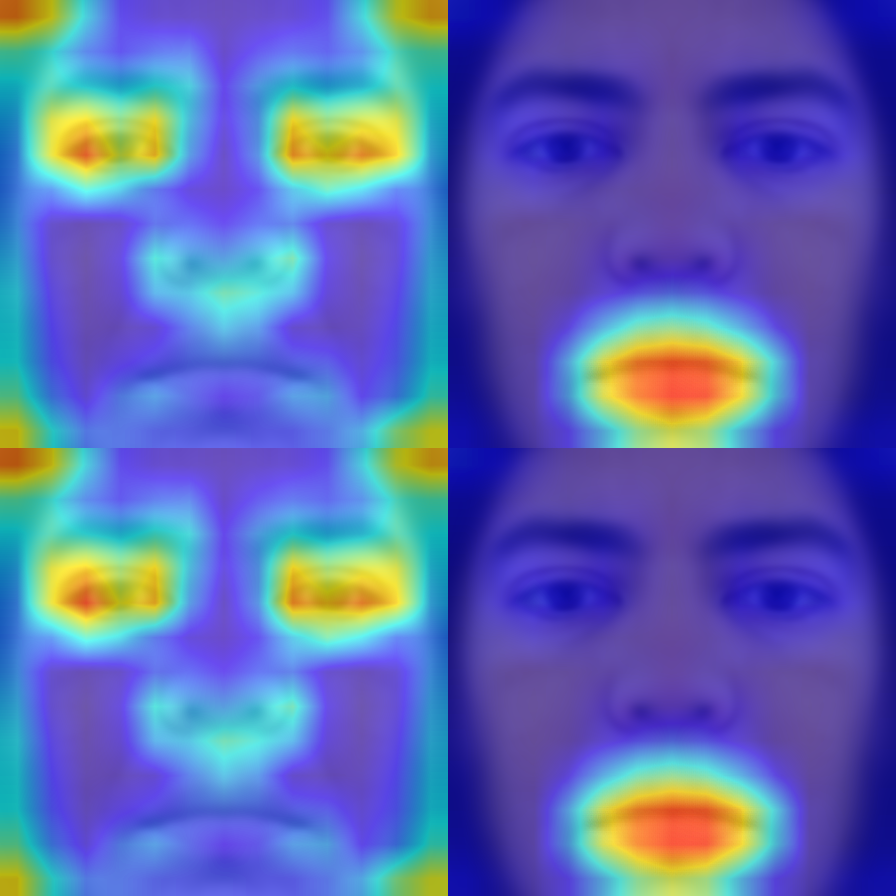}
            \put(3, 52){\tiny\color{white}{before, $pitch$}}
            \put(56, 52){\tiny\color{white}{after, $pitch$}}
            \put(6, 2){\tiny\color{white}{before, $yaw$}}
            \put(59, 2){\tiny\color{white}{after, $yaw$}}
         \end{overpic}
      \end{minipage}
      \label{fig:attention_mouth}
   }
   \subfigure[Others]{
      \begin{minipage}[b]{0.3\linewidth}
         \begin{overpic}[width=1\textwidth]{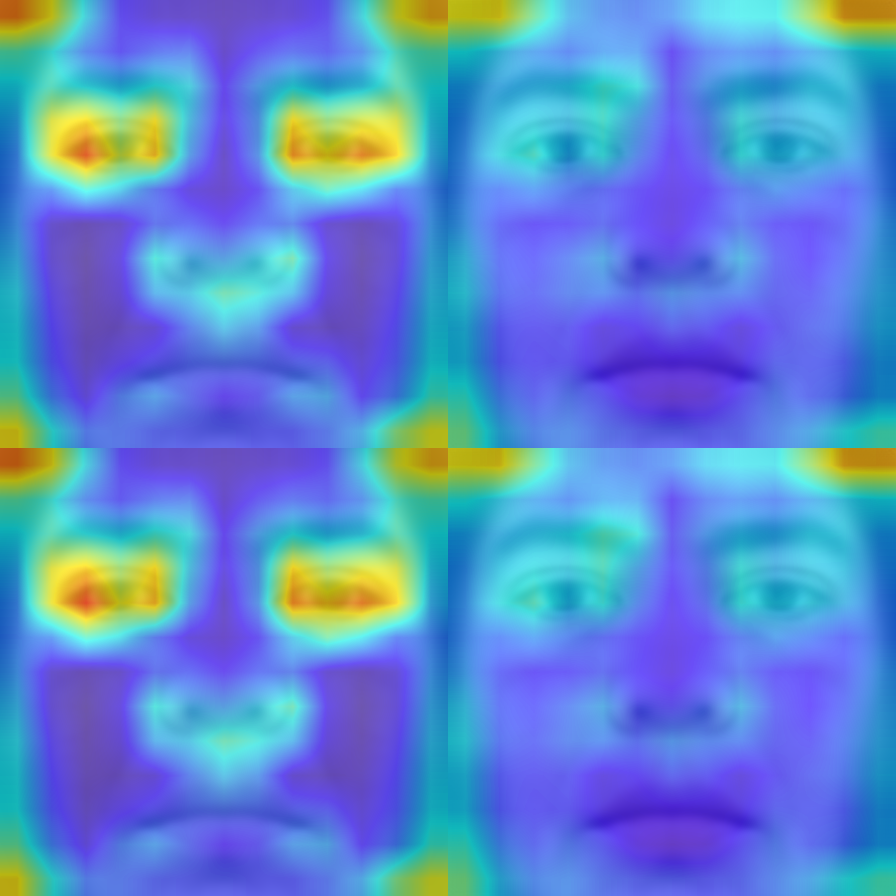}
            \put(3, 52){\tiny\color{white}{before, $pitch$}}
            \put(56, 52){\tiny\color{white}{after, $pitch$}}
            \put(6, 2){\tiny\color{white}{before, $yaw$}}
            \put(59, 2){\tiny\color{white}{after, $yaw$}}
         \end{overpic}
      \end{minipage}
      \label{fig:attention_others}
   }
   \subfigure[Eyes+Nose]{
      \begin{minipage}[b]{0.3\linewidth}
         \begin{overpic}[width=1\textwidth]{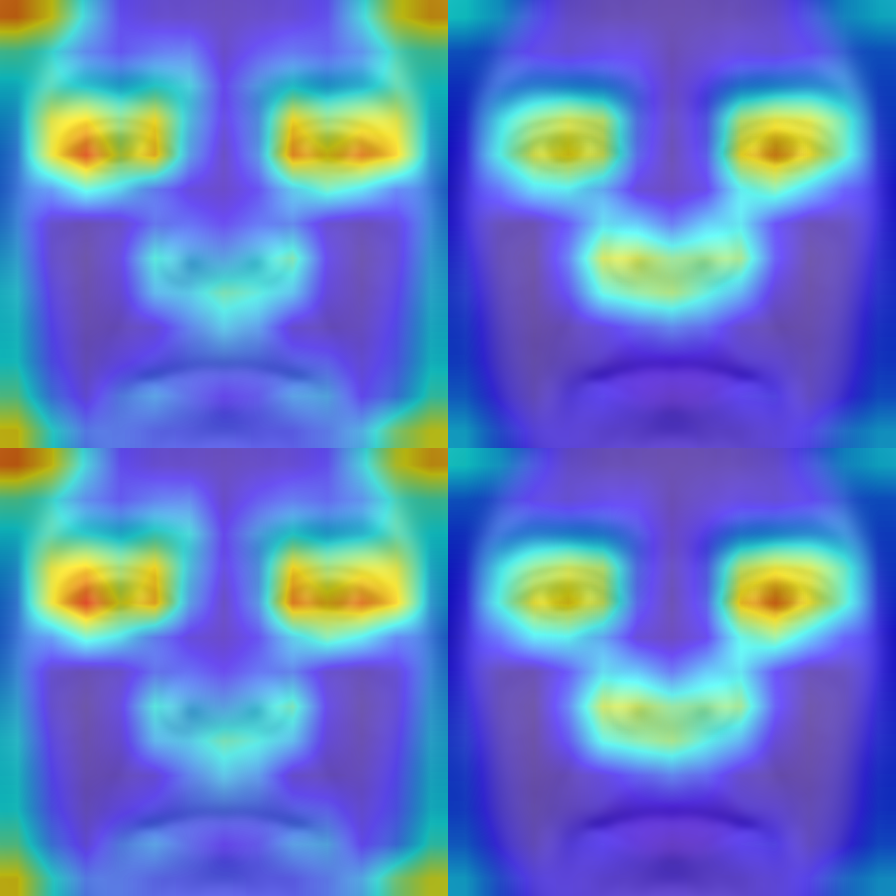}
            \put(3, 52){\tiny\color{white}{before, $pitch$}}
            \put(56, 52){\tiny\color{white}{after, $pitch$}}
            \put(6, 2){\tiny\color{white}{before, $yaw$}}
            \put(59, 2){\tiny\color{white}{after, $yaw$}}
         \end{overpic}
      \end{minipage}
      \label{fig:attention_eyes_nose}
   }
   \subfigure[Nose+Mouth]{
      \begin{minipage}[b]{0.3\linewidth}
         \begin{overpic}[width=1\textwidth]{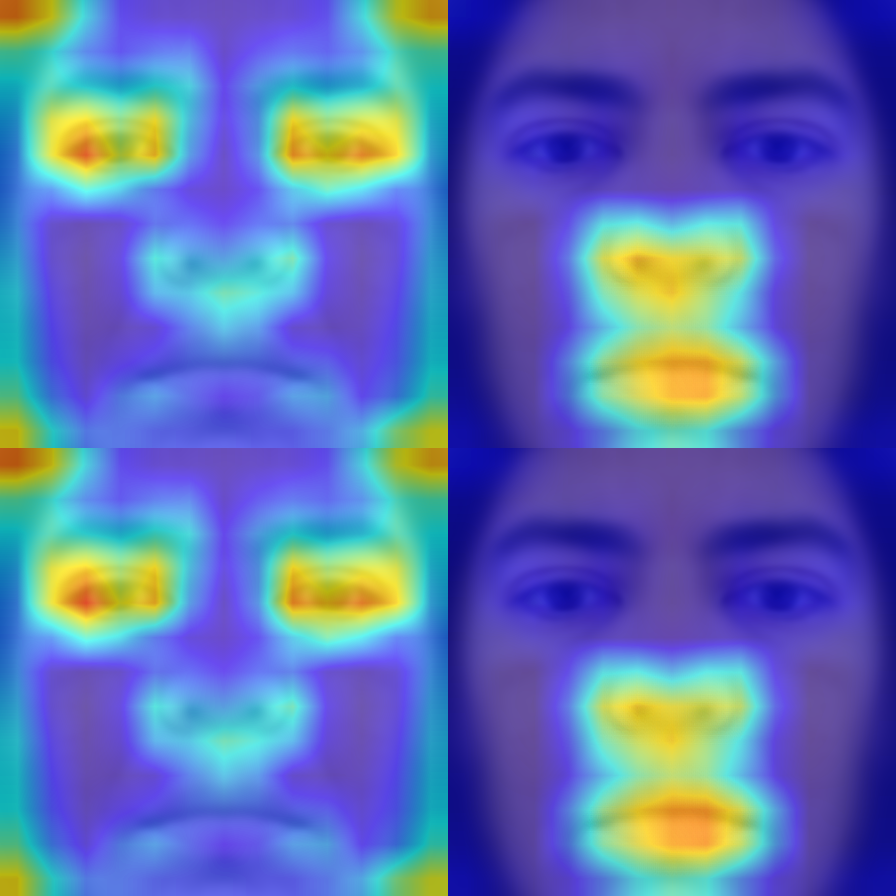}
            \put(3, 52){\tiny\color{white}{before, $pitch$}}
            \put(56, 52){\tiny\color{white}{after, $pitch$}}
            \put(6, 2){\tiny\color{white}{before, $yaw$}}
            \put(59, 2){\tiny\color{white}{after, $yaw$}}
         \end{overpic}
      \end{minipage}
      \label{fig:attention_nose_mouth}
   }
   \subfigure[Eyes+Mouth]{
      \begin{minipage}[b]{0.3\linewidth}
         \begin{overpic}[width=1\textwidth]{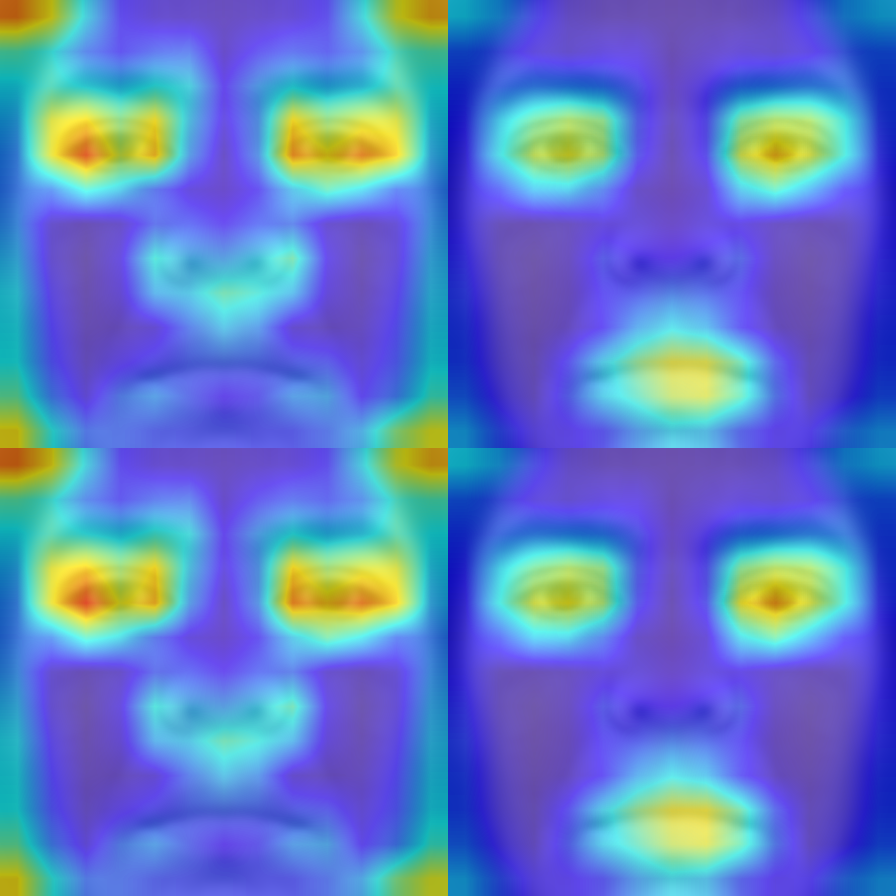}
            \put(3, 52){\tiny\color{white}{before, $pitch$}}
            \put(56, 52){\tiny\color{white}{after, $pitch$}}
            \put(6, 2){\tiny\color{white}{before, $yaw$}}
            \put(59, 2){\tiny\color{white}{after, $yaw$}}
         \end{overpic}
      \end{minipage}
      \label{fig:attention_eyes_mouth}
   }
   \subfigure[Eyes+Nose+Mouth]{
      \begin{minipage}[b]{0.3\linewidth}
         \begin{overpic}[width=1\textwidth]{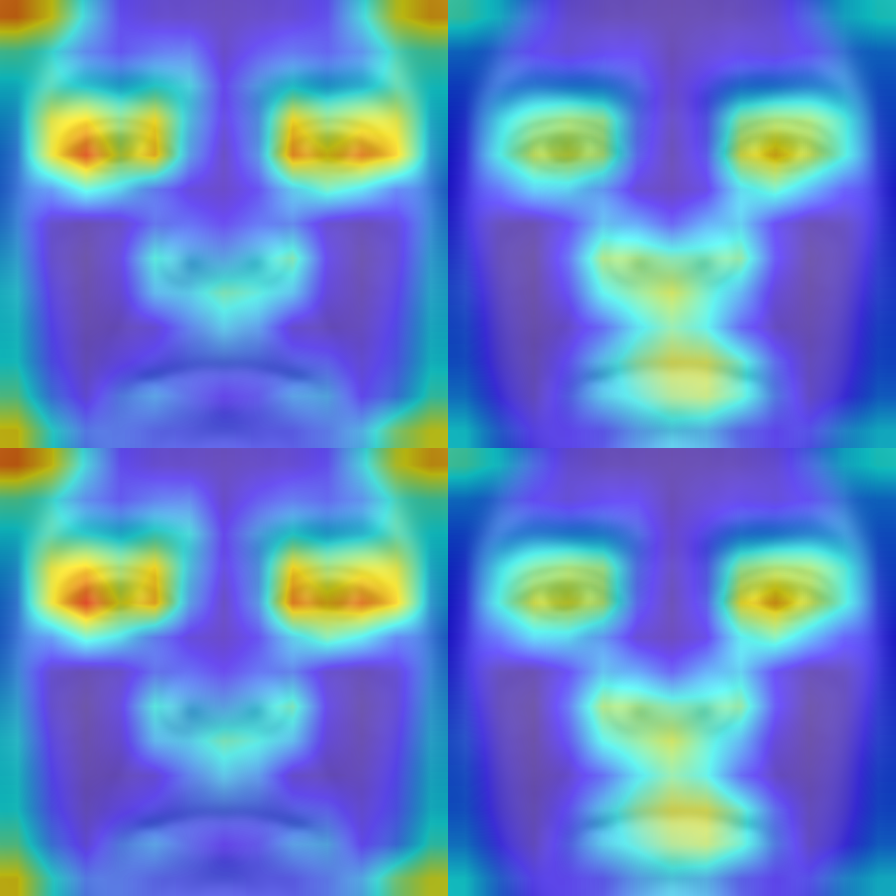}
            \put(3, 52){\tiny\color{white}{before, $pitch$}}
            \put(56, 52){\tiny\color{white}{after, $pitch$}}
            \put(6, 2){\tiny\color{white}{before, $yaw$}}
            \put(59, 2){\tiny\color{white}{after, $yaw$}}
         \end{overpic}
      \end{minipage}
      \label{fig:attention_eyes_nose_mouth}
   }
   \subfigure[E+N+M+O]{
      \begin{minipage}[b]{0.3\linewidth}
         \begin{overpic}[width=1\textwidth]{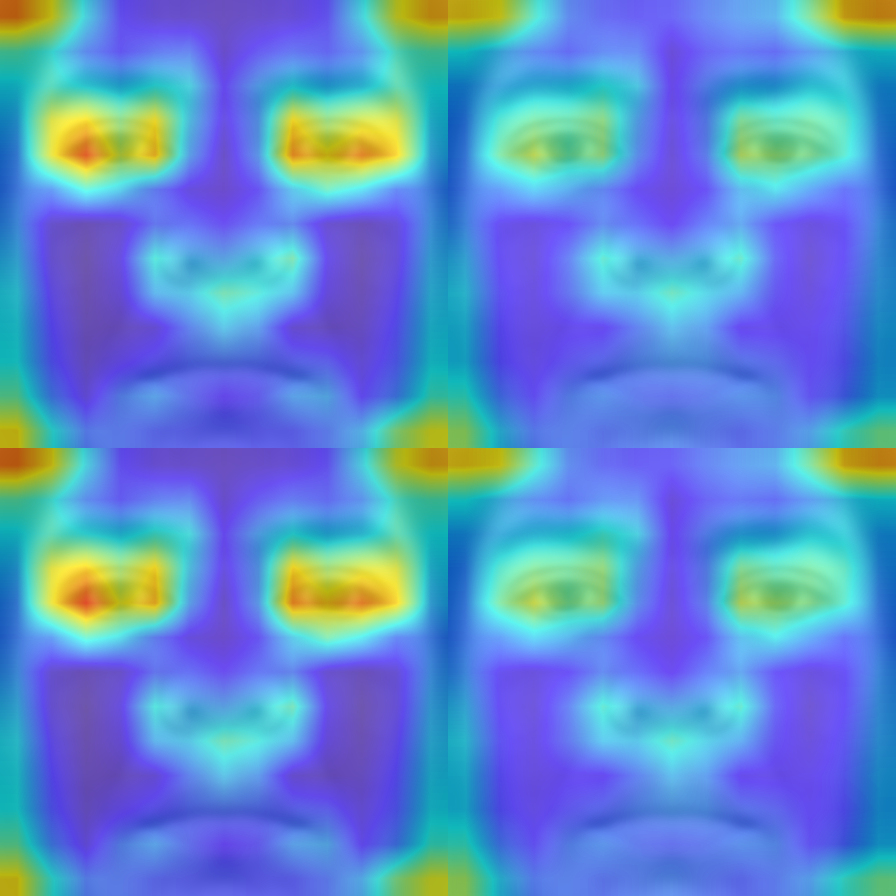}
            \put(3, 52){\tiny\color{white}{before, $pitch$}}
            \put(56, 52){\tiny\color{white}{after, $pitch$}}
            \put(6, 2){\tiny\color{white}{before, $yaw$}}
            \put(59, 2){\tiny\color{white}{after, $yaw$}}
         \end{overpic}
      \end{minipage}
      \label{fig:full_face}
   }
   \caption{Visualization of the attention area before and after attacking different parts of the face. For each subfigure, ``before'' means before attack, ``after'' means after attack.}
\label{fig:attention}
\end{figure}
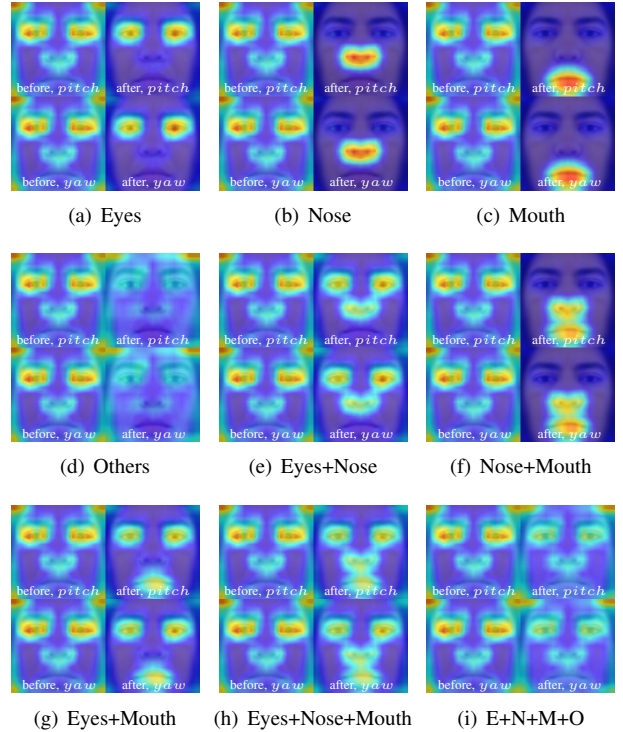

While the attacks on Full-Face performed above are only for the full face, it is necessary to know the effect of attacking different face regions. The face regions are shown in Figure \ref{fig:face_parts}, including ``Eyes'', ``Nose'', ``Mouth'' and ``Others'' (the rest parts of the face). Here we conduct the attack experiment on Full-Face model and show the results when $\epsilon=32, \alpha=0.25$. This is because these hyper-parameters generate the typical results. Table \ref{table:face_parts} displays the result. We found that the attack on ``Eyes'' achieves the lowest angular error. This can be explained that before attack, ``Eyes'' is the most focused region in Figure \ref{fig:attention}, which indicates that ``Eyes'' is the most vulnerable part for attacks. Interestingly, attack on ``Others'' performs relatively poor although its mean angle error is low enough.

To investigate the reason behind vulnerability and how the attack influences the model, we visualized the attention patterns using Score-CAM \cite{wang2020score} on Full-face model, as shown in Figure \ref{fig:attention}. It can be observed that ``Others'' does not attract much attention, and Figure \ref{fig:attention_others} illustrates that the attention pattern is diluted after the attack. It can be concluded that the attention pattern changes even though we attack on such unimportant part, thus it validates the model vulnerability. We also observe that the system achieves large angular error when we only attack on either ``Nose'' or ``Mouth''. Figure \ref{fig:attention_nose} and \ref{fig:attention_mouth} shows that the attention area is only ``Nose'' or ``Mouth'' after attack. If both ``Nose'' and ``Mouth'' are attacked, much lower angular error can be achieved and the attention area also transfers to both ``Nose'' and ``Mouth'' after attack. But this angular error is still higher than that performed by attack on some other parts. Note that all the attention areas are not on ``Eyes'', which leads to such result. If we attack on both ``Eyes'' and ``Nose'', or both ``Eyes'' and ``Mouth'', the mean angular error becomes much lower compared to the attack on ``nose'' or ``mouth'' only. In this case, the attention area is on both ``Eyes'' and some other parts. 

Based on the attention patterns, we found that if the attention of ``Eyes'' is strong as in Figure \ref{fig:attention_eyes}, \ref{fig:attention_eyes_nose}, \ref{fig:attention_eyes_mouth} and \ref{fig:attention_eyes_nose_mouth}, the system achieves very low angular error. If the attention of ``Eyes'' exists but it is not strong, \eg, Figure \ref{fig:attention_others} and \ref{fig:full_face}, low angular error can also be obtained but higher than the previous case; if the attention of ``Eyes'' almost does not exist, we can only achieve large angular error as shown in Figure \ref{fig:attention_nose}, \ref{fig:attention_mouth} and \ref{fig:attention_nose_mouth}.

\begin{table}[!hbt]
\caption{Performance of attacking on different parts of the face. It shows the mean $\pm$ standard deviation of the angular error across all inputs and targets.}
\label{table:face_parts}
\centering
\normalsize{{
   \begin{tabular}{cccc|c}
      \toprule
      \textbf{Face parts} & \textbf{Angular error} \\
      \midrule
      \underline{Eyes} & \textbf{0.283}$\pm$0.374 \\
      Nose & 30.33$\pm$12.53 \\
      Mouth & 26.61$\pm$18.05 \\
      Others & 1.047$\pm$0.651 \\
      \midrule
      Nose+Mouth & 4.402$\pm$7.710 \\
      \underline{Eyes}+Nose & 0.320$\pm$0.372 \\
      \underline{Eyes}+Mouth & 0.341$\pm$0.383 \\
      \underline{Eyes}+Nose+Mouth & 0.412$\pm$0.422 \\
      \midrule
      \underline{Eyes}+Nose+Mouth+Others & 1.421$\pm$0.614 \\
      \bottomrule
   \end{tabular}}}
\end{table}

\subsubsection{Smoothness}
\label{section:smoothness}

From Fig. \ref{fig:epsilon}, We observed that sometimes the adversarial perturbation using large $\epsilon$ is obvious enough to be distinguished. From the perspective of the attacker, this makes the application of attacks limited. To make the perturbation more smooth, we introduce the total variation loss \cite{mahendran2015understanding}. Here we modify the objective function (\ref{eq:mean_angle_loss}) to:
\begin{equation}
   \begin{split}
      \mathcal{L}(\mathcal{G}(\widetilde{\mathbf{x}}, \dots), \mathbf{t}) = \cos^{-1}{\left( \frac{\mathbf{t} \cdot \mathcal{G}(\widetilde{\mathbf{x}}, \dots)}{ \| \mathbf{t}\| \cdot \| \mathcal{G}(\widetilde{\mathbf{x}}, \dots) \|} \right)} +\lambda_{TV}\text{TV}(\widetilde{\mathbf{x}})
      \label{eq:smooth_loss}
   \end{split}
\end{equation}
where $\lambda_{TV}$ controls the weights between the angular error and the total variation loss, $\text{TV}(\widetilde{\mathbf{x}})$ is the total variation loss, defined as:
\begin{equation}
   \begin{split}
      &\text{TV}(\mathbf{x})=\frac{\sum\limits_{i=1}^{h-1}\sum\limits_{j=1}^{w-1}{\left[(\mathbf{x}_{i,j+1}-\mathbf{x}_{i,j})^2+(\mathbf{x}_{i+1,j}-\mathbf{x}_{i,j})^2\right]}}{(h-1)(w-1)}.
      \label{eq:tv_loss}
   \end{split}
\end{equation}

\begin{figure}[t]
   \centering
   \includegraphics[width=\linewidth]{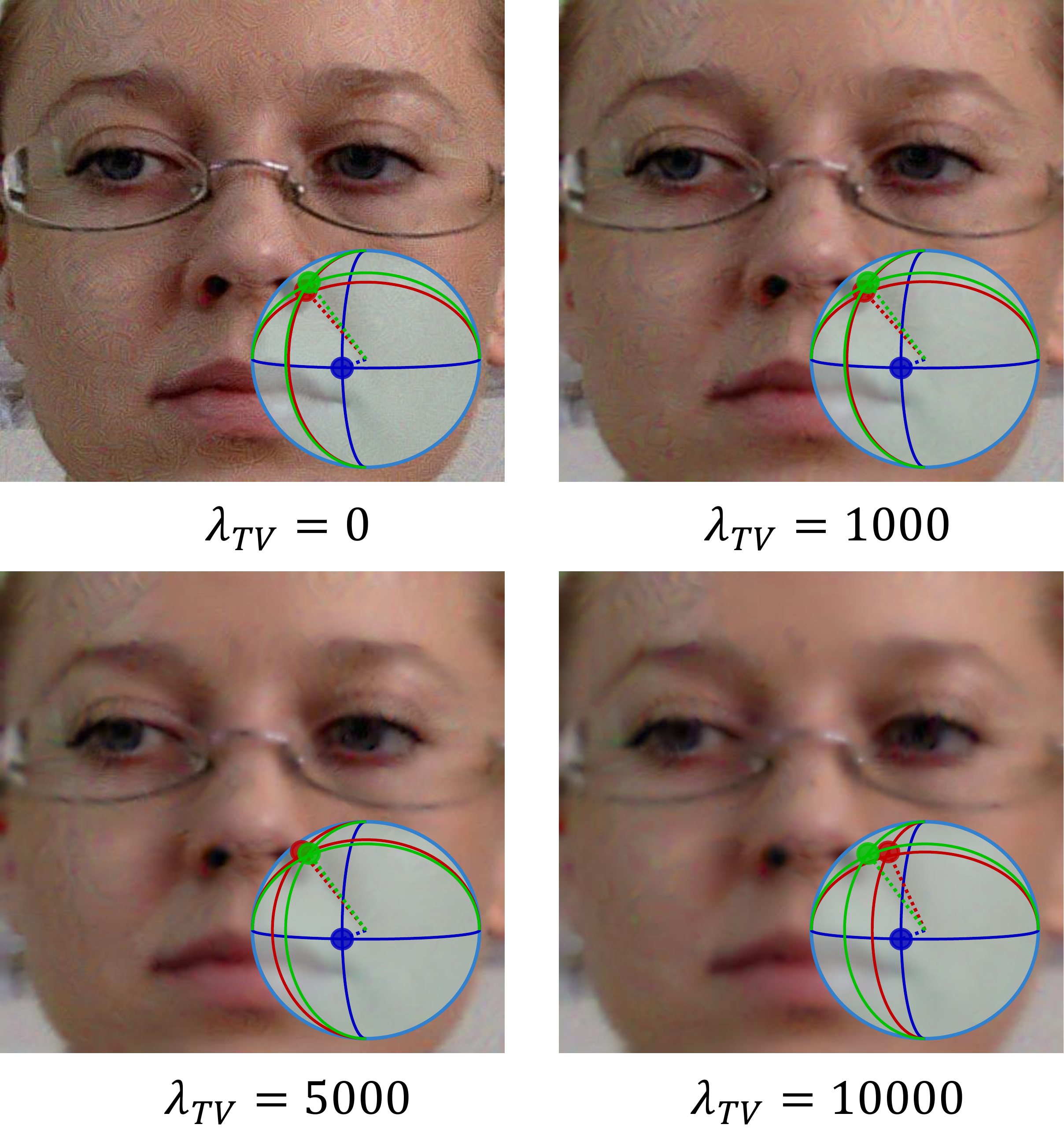}
   \caption{Example of perturbed images under different $\lambda_{TV}$. Blue arrow is the estimated gaze before attack, red arrow is the estimated gaze after attack, and green arrow is the target gaze direction.}
   
\label{fig:smoothness}
\end{figure}

Figure \ref{fig:smoothness} illustrates the result of smoothness, we observe that when we increase $\lambda_{TV}$, the perturbed image becomes smoother. However, the attacked gaze direction only slightly changes, this validates the vulnerability of the gaze estimation models, even indistinguishable noise could perturb the network prediction.

\section{Study on patch-based adversarial attack}
\label{section:patch-attack}

\subsection{Method}

Although the pixel-based method is effective for attack on appearance-based gaze estimation as discussed in Section \ref{section:pixel-attack}, it is difficult to make such method physically realizable. Due to this reason, we draw on the idea of adversarial patch \cite{brown2017adversarial} and place a patch on the face common to all the input images (including the content and the position of the patch), as shown in Fig. \ref{fig:patch}, instead of adding adversarial perturbation directly on the face. In order to make the face detectable, we do not place the patch on the landmarks of the face. 

First, we define the adversarial example as:
\begin{equation}
   \begin{split}
      \widetilde{\mathbf{x}}= \mathbf{x} \odot (1-\mathbf{m})+\mathbf{p} \odot \mathbf{m}
   \end{split}
\end{equation}
where $\odot$ is element-wise multiplication, $\mathbf{p}$ is the patch whose size is the same as $\mathbf{x}$, $\mathbf{m}_{i,j}=1$ means $(i,j)$ belongs to the patch and $\mathbf{m}_{i,j}=0$ means $(i,j)$ doesn't belong to the patch. Obviously, $\mathbf{m}$ defines the region of the patch.

The patch-based method is also based on the idea of Basic Iterative Method (BIM) method \cite{kurakin2016adversarial}:
\begin{equation}
   \begin{split}
      \widetilde{\mathbf{x}}_k &= \mathbf{x} \odot (1-\mathbf{m})+\mathbf{p}_{k} \odot \mathbf{m},\\
      \mathbf{p}_{k} &= \mathbf{p}_{k-1} - \alpha \; \text{sign} \bigl( \nabla_\mathbf{p} \mathcal{L}(\mathcal{G}(\widetilde{\mathbf{x}}_{k-1}, \dots), \mathbf{t})  \bigr), \label{eq:bim_patch}
   \end{split}
\end{equation}
where $k\in [1,N]$ is the index of the current iteration. If the current $\mathbf{x}$ is the first image to input, $\mathbf{p}_0$ is the randomly initialized image. Otherwise, $\mathbf{p}_0$ is the $\mathbf{p}_k$ related to the minimum angular error of the previous image. Note that all the images used to generate the patch will be iterated for $num\_epochs$ epochs and we have removed the $\| \widetilde{\mathbf{x}}-\mathbf{x}\| \le \epsilon$ constraint described in Eq. (\ref{eq:optimization}).

\subsection{Characterization}
\label{section:patch-performance-analysis}

We choose Full-Face model to evaluate the effectiveness of patch-based attack. Here we show the results when $\alpha=1$ and $num\_epochs=5$ since these parameters generate the typical results. The mask $\mathbf{m}$ of the patch is a circle with center $(271,358)$ and radius $48$. In this experiment, 10\% of the data uniformly sampled from the test set is used to attack, and the full test set is used to evaluate the effects of attack. The total loss is defined as:
\begin{equation}
   \begin{split}
      \mathcal{L}(\mathcal{G}(\widetilde{\mathbf{x}}, \dots), \mathbf{t}) =&\cos^{-1}{\left( \frac{\mathbf{t} \cdot \mathcal{G}(\widetilde{\mathbf{x}}, \dots)}{ \| \mathbf{t}\| \cdot \| \mathcal{G}(\widetilde{\mathbf{x}}, \dots) \|} \right)}
      \\ +&\lambda_{TV}\text{TV}(\mathbf{p} \odot \mathbf{m}),
      \label{eq:loss_patch}
   \end{split}
\end{equation}

\begin{figure}[t]
   \centering
   \subfigure[$\lambda_{TV}=0$]{
      \begin{minipage}[b]{0.3\linewidth}
         \includegraphics[width=1.0\linewidth]{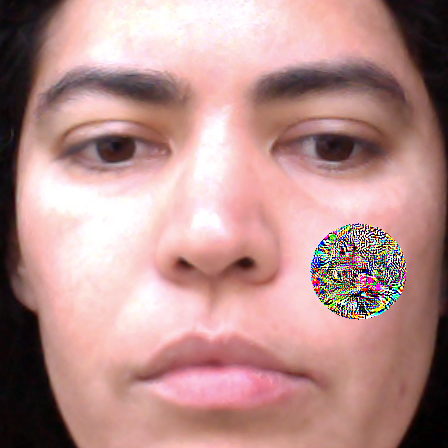}
      \end{minipage}
   }
   \subfigure[$\lambda_{TV}=1000$]{
      \begin{minipage}[b]{0.3\linewidth}
         \includegraphics[width=1.0\linewidth]{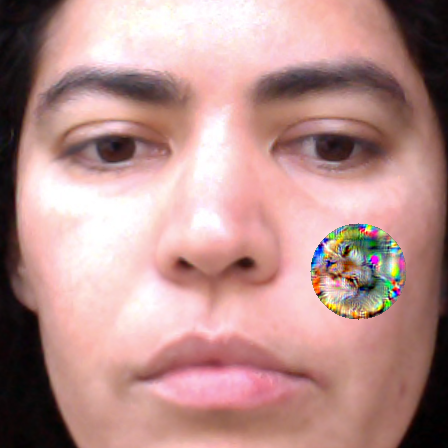}
      \end{minipage}
   }
   \subfigure[$\lambda_{TV}=5000$]{
      \begin{minipage}[b]{0.3\linewidth}
         \includegraphics[width=1.0\linewidth]{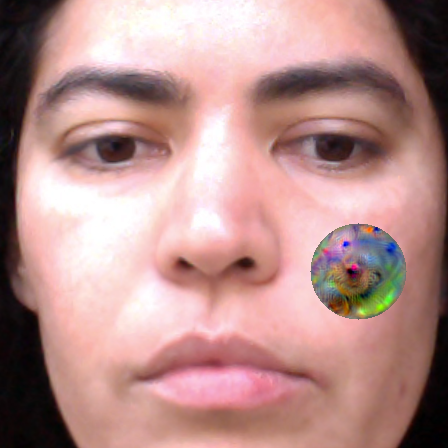}
      \end{minipage}
   }
   \caption{Visualization of the adversarial examples of the patch-based method. We used Person \#0 in MPIIFaceGaze and the target Q2.}
\label{fig:patch}
\end{figure}

Experimental results are shown in Table \ref{table:patch} and the visualization of the attack is displayed in Fig. \ref{fig:patch}. We observed that the patch-based method is also effective to attack the model. But the smoothness of the patch comes at the expense of the attack effect. The method is easy to be implemented in real world settings because we can easily print and stick the patch onto the specific region of the face, then take photos and input the photo to the model. However, the result of the real world setting may not be satisfactory due to the limitation of training on only one dataset and the color difference between the printed image and the image displayed on the screen.

\begin{table}[!hbt]
\caption{Performance of the patch-based method under different $\lambda_{TV}$. It shows the mean $\pm$ standard deviation of the angular error across all inputs and targets.}
\label{table:patch}
\centering
\normalsize{{
   \begin{tabular}{c|c|c|c}
      \toprule
      $\lambda_{TV}$ & 0 & 1000 & 5000 \\
      \midrule
      \textbf{Error} & 10.13$\pm$5.07 & 18.03$\pm$9.76 & 38.35$\pm$17.52 \\
      \bottomrule
   \end{tabular}}}
\end{table}

\section{Study on defense}
\label{section:study-on-defense}

We have discussed the vulnerability of models on the attack task in last two sections. How well will these models perform on the defense task remains to be explored. The aim of defense is to make the estimated gaze directions farther from the target gaze directions but closer to the ground truth.

\subsection{Method}

Adversarial training is commonly used as a method for defending the adversarial attacks \cite{goodfellow2014explaining}. Madry \etal \cite{madry2017towards} proposed the adversarial training method against project gradient descent attacks, but the computational cost is high. Toward this problem, Shafahi \etal \cite{shafahi2019adversarial} introduced free adversarial training to reduce the computational cost. Here, we follow the free adversarial training \cite{shafahi2019adversarial} and the idea of stability training \cite{zheng2016improving} for defense. We use the gradient of mean angular error to generate the adversarial perturbations. The objective function of training is
\begin{equation}
   \begin{split}
      &\mathcal{L}_{model}(\mathcal{G}(\mathbf{x}, \dots), \mathbf{y})+\lambda_{adv}\mathcal{L}_{model}(\mathcal{G}(\mathbf{x}, \dots), \mathcal{G}(\widetilde{\mathbf{x}}, \dots)).
      \label{eq:training_loss_adv}
   \end{split}
\end{equation}
where $\mathcal{L}_{model}$ is the original training loss. This objective function takes into account the training of original examples, and the goal is to make the adversarial examples and the original examples as close as possible. $\lambda_{adv}$ balances these two goals. In free adversarial training \cite{shafahi2019adversarial}, given the original number of training epochs $N_{ep}$, free adversarial training runs for $N_{ep}/m$ epochs, and in each epoch, the adversarial example generation will run for $m$ consecutive iterations. $\epsilon$ is the $\| \widetilde{\mathbf{x}}-\mathbf{x} \| \le \epsilon$ constraint of adversarial examples for free adversarial training and $\alpha$ is the stride of one iteration generating adversarial examples.

\subsection{Performance analysis}

We test the defense method on Full-Face, Gaze-Net and RT-GENE, except for CA-Net. This is due to the superior performance of the CA-Net in the attack task. Table \ref{table:loss_after_adv} displays the training information with and without defense. Note that we choose the hyper-parameter that attempts to maximize the effectiveness of defense for each model, but we failed to achieve the best defensive effect.

\begin{table}[!hbt]
\caption{Training information of methods with and without defense. ``w/o defense'' and ``w/ defense'' are the mean angular errors across all the inputs and targets. $m$, $\epsilon$, $\alpha$ and $\lambda_{adv}$ are the hyper-parameters of adversarial training.}
\label{table:loss_after_adv}
\centering
\normalsize{\resizebox{\linewidth}{!}{
   \begin{tabular}{c|cc|cccc}
      \toprule
      \multirow{2}[2]{*}{\shortstack{\textbf{Dataset}+\textbf{Method}}} & \multirow{2}[2]{*}{\shortstack{\textbf{w/o defense}}} & \multirow{2}[2]{*}{\shortstack{\textbf{w/ defense}}} & \multicolumn{4}{c}{\textbf{Hyper-parameter}} \\ 
      & & & $m$ & $\epsilon$ & $\alpha$ & $\lambda_{adv}$ \\
      \midrule
      \shortstack{MPIIFaceGaze\cite{zhang2017s}\\+Full-Face\cite{zhang2017s}} & 2.455 & 2.672 & 5 & 64 & 0.25 & 0.01 \\
      \midrule
      \shortstack{MPIIGaze\cite{DBLP:journals/corr/abs-1711-09017}\\+Gaze-Net\cite{DBLP:journals/corr/abs-1711-09017}} & 4.565 & 3.905 & 5 & 128 & 1.0 & 0.05 \\
      \midrule
      \shortstack{RT-GENE \cite{FischerECCV2018}} & 10.23 & 14.23 & 1 & 64 & 0.25 & 0.005 \\
      \midrule
      \shortstack{RT-GENE \cite{FischerECCV2018}\\(Augmented)} & 10.87 & 15.54 & 1 & 64 & 0.125 & 0.005 \\
      \bottomrule
   \end{tabular}}}
\end{table}

For each method, due to the computational cost, only one fold of the test set (Person \#0 for MPIIFaceGaze and MPIIGaze, Person \#1 for EYEDIAP, Fold \#0 for RT-GENE) is used to attack, and the other folds are used to train the model, \ie, leave-one-person-out strategy. Here we empirically choose $\alpha=0.125$. Table \ref{table:attack_adv}, \ref{table:gt_before}, \ref{table:gt_after} and Figure \ref{table:attack_adv} reports the attack effect after defense. 
In general, it is difficult to predict the target gaze direction for these models. However, for Full-Face and RT-GENE, it is still relatively easy to be attacked at certain targets.

\begin{table}[!hbt]
    \caption{Evaluation results of defense. Except for $\epsilon$, all the other numbers are mean angle error between the target and estimated gaze direction of all the inputs in one fold plus minus the standard deviation of them.}
    \label{table:attack_adv}
    \centering
    \normalsize{\resizebox{\linewidth}{!}{
       \begin{tabular}{c|c|ccccc}
          \toprule
          \multirow{2}[2]{*}{\shortstack{\textbf{Dataset}\\+ \textbf{Method}}} & \multirow{2}[2]{*}{$\epsilon$} & \multicolumn{5}{c}{
             \textbf{Target}
          } \\
          & & Q1 & Q2 & Q3 & Q4 & \textbf{Average} \\
          \midrule
          \multirow{3}[2]{*}{\shortstack{MPIIFaceGaze\cite{zhang2017s}\\+ Full-Face\cite{zhang2017s}}} 
             & 16 & 20.8$\pm$2.5 & 17.7$\pm$2.2 & 2.1$\pm$1.7 & 1.0$\pm$0.8 & \textbf{10.4$\pm$9.1} \\
             & 32 & 13.8$\pm$2.2 & 11.7$\pm$1.9 & 0.3$\pm$1.0 & 0.4$\pm$0.5 & \textbf{6.6$\pm$6.4} \\
             & 64 & 6.9$\pm$2.2  & 5.8$\pm$1.8  & 0.1$\pm$0.5  & 0.3$\pm$0.3 & \textbf{3.3$\pm$3.4} \\
          \midrule
          \multirow{3}[2]{*}{\shortstack{MPIIGaze\cite{DBLP:journals/corr/abs-1711-09017}\\+ Gaze-Net\cite{DBLP:journals/corr/abs-1711-09017}}} 
             & 16 & 43.6$\pm$3.9 & 44.5$\pm$4.8 & 30.2$\pm$6.0 & 29.3$\pm$5.8 & \textbf{36.9$\pm$8.8} \\
             & 32 & 35.0$\pm$3.8 & 34.8$\pm$4.6 & 20.2$\pm$5.3 & 19.7$\pm$5.5 & \textbf{27.4$\pm$8.9} \\
             & 64 & 28.3$\pm$4.1 & 26.9$\pm$4.9 & 13.1$\pm$4.7 & 12.9$\pm$5.3 & \textbf{20.3$\pm$8.7} \\
          \midrule
          \multirow{3}[2]{*}{\shortstack{RT-GENE \cite{FischerECCV2018}}} 
             & 16 & 15.9$\pm$7.6  & 22.7$\pm$12.2 & 15.4$\pm$11.6 & 3.2$\pm$4.8 & \textbf{14.3$\pm$11.9} \\
             & 32 & 7.8$\pm$6.2   & 12.2$\pm$10.4 & 8.3$\pm$9.4  & 1.7$\pm$3.4 & \textbf{7.5$\pm$8.7} \\
             & 64 & 4.4$\pm$5.1   & 6.7$\pm$7.6   & 5.6$\pm$8.1  & 1.2$\pm$2.7 & \textbf{4.5$\pm$6.6} \\
          \midrule
          \multirow{3}[2]{*}{\shortstack{RT-GENE \cite{FischerECCV2018}\\(Augmented)}} 
             & 16 & 41.0$\pm$5.8  & 40.1$\pm$12.0 & 40.6$\pm$6.5 & 43.1$\pm$4.4 & \textbf{41.2$\pm$7.8} \\
             & 32 & 35.0$\pm$7.5  & 31.2$\pm$14.7 & 34.6$\pm$7.3 & 38.8$\pm$5.2 & \textbf{34.9$\pm$9.8} \\
             & 64 & 26.1$\pm$9.8  & 23.4$\pm$15.4 & 27.8$\pm$8.4 & 34.6$\pm$6.7 & \textbf{28.0$\pm$11.4} \\
          \bottomrule
       \end{tabular}}}
    \end{table}

\begin{table}[!hbt]
\caption{The mean angular error between the ground truth and the estimated gaze direction after attack \textbf{without defense}. The mean angular error of all the inputs in one fold is displayed.}
\label{table:gt_before}
\centering
\normalsize{\resizebox{\linewidth}{!}{
    \begin{tabular}{c|c|ccccc}
        \toprule
        \multirow{2}[2]{*}{\shortstack{\textbf{Dataset}\\+ \textbf{Method}}} & \multirow{2}[2]{*}{$\epsilon$} & \multicolumn{5}{c}{
            \textbf{Target}
        } \\
        & & Q1 & Q2 & Q3 & Q4 & \textbf{Average} \\
        \midrule
        \multirow{3}[2]{*}{\shortstack{MPIIFaceGaze\\+ Full-Face\cite{zhang2017s}}} 
            & 16 & 68.9$\pm$7.7 & 69.1$\pm$7.7 & 52.3$\pm$8.2 & 52.1$\pm$8.3 & \textbf{60.6$\pm$11.6} \\
            & 32 & 68.9$\pm$7.7 & 69.1$\pm$7.7 & 52.3$\pm$8.2 & 52.2$\pm$8.2 & \textbf{60.6$\pm$11.6} \\
            & 64 & 69.0$\pm$7.7 & 69.1$\pm$7.7 & 52.3$\pm$8.2 & 52.2$\pm$8.2 & \textbf{60.7$\pm$11.6} \\
        \midrule
        \multirow{3}[2]{*}{\shortstack{MPIIGaze\\+ Gaze-Net\cite{zhang15_cvpr,DBLP:journals/corr/abs-1711-09017}}} 
            & 16 & 35.9$\pm$8.2 & 36.9$\pm$8.1 & 30.5$\pm$9.0 & 34.5$\pm$8.7 & \textbf{34.5$\pm$8.9} \\
            & 32 & 51.3$\pm$8.2 & 53.9$\pm$8.2 & 44.1$\pm$8.6 & 47.9$\pm$7.8 & \textbf{49.3$\pm$9.0} \\
            & 64 & 64.7$\pm$7.2 & 66.5$\pm$6.6 & 51.8$\pm$6.6 & 52.5$\pm$6.3 & \textbf{58.9$\pm$9.5} \\
        \midrule
        \multirow{3}[2]{*}{\shortstack{RT-GENE \cite{FischerECCV2018}}} 
            & 16 & 55.3$\pm$11.7 & 60.4$\pm$12.8 & 60.2$\pm$11.9 & 56.4$\pm$11.7 & \textbf{58.1$\pm$12.3} \\
            & 32 & 57.9$\pm$11.8 & 62.6$\pm$12.2 & 60.3$\pm$12.0 & 56.5$\pm$11.7 & \textbf{59.3$\pm$12.2} \\
            & 64 & 59.3$\pm$11.9 & 63.9$\pm$12.0 & 60.3$\pm$12.0 & 56.6$\pm$11.7 & \textbf{60.0$\pm$12.2} \\
        \midrule
        \multirow{3}[2]{*}{\shortstack{RT-GENE \cite{FischerECCV2018}\\(Augmented)}} 
            & 16 & 58.6$\pm$11.7 & 61.2$\pm$11.9 & 55.3$\pm$11.6 & 55.7$\pm$12.1 & \textbf{57.7$\pm$12.1} \\
            & 32 & 61.0$\pm$11.6 & 63.1$\pm$11.7 & 57.0$\pm$11.6 & 56.3$\pm$11.8 & \textbf{59.3$\pm$12.0} \\
            & 64 & 61.5$\pm$11.6 & 64.1$\pm$11.6 & 57.9$\pm$11.6 & 56.5$\pm$11.7 & \textbf{60.0$\pm$12.0} \\
        \bottomrule
    \end{tabular}}}
\end{table}

\begin{table}[!hbt]
\caption{The mean angular error between the ground truth and the estimated gaze direction after attack \textbf{with defense}. The mean angular error of all the inputs in one fold is displayed.}
\label{table:gt_after}
\centering
\normalsize{\resizebox{\linewidth}{!}{
    \begin{tabular}{c|c|ccccc}
        \toprule
        \multirow{2}[2]{*}{\shortstack{\textbf{Dataset}\\+ \textbf{Method}}} & \multirow{2}[2]{*}{$\epsilon$} & \multicolumn{5}{c}{
            \textbf{Target}
        } \\
        & & Q1 & Q2 & Q3 & Q4 & \textbf{Average} \\
        \midrule
        \multirow{3}[2]{*}{\shortstack{MPIIFaceGaze\\+ Full-Face\cite{zhang2017s}}} 
            & 16 & 49.1$\pm$7.9 & 51.9$\pm$8.2 & 50.4$\pm$7.6 & 51.4$\pm$8.0 & \textbf{50.7$\pm$8.0} \\
            & 32 & 55.7$\pm$7.9 & 57.7$\pm$8.0 & 52.1$\pm$8.1 & 52.0$\pm$8.2 & \textbf{54.4$\pm$8.4} \\
            & 64 & 62.4$\pm$8.0 & 63.5$\pm$7.9 & 52.2$\pm$8.1 & 52.1$\pm$8.2 & \textbf{57.5$\pm$9.7} \\
        \midrule
        \multirow{3}[2]{*}{\shortstack{MPIIGaze\\+ Gaze-Net\cite{zhang15_cvpr,DBLP:journals/corr/abs-1711-09017}}} 
            & 16 & 25.9$\pm$7.1 & 25.1$\pm$7.4 & 23.8$\pm$8.5 & 24.6$\pm$8.1 & \textbf{24.8$\pm$7.8} \\
            & 32 & 34.0$\pm$7.1 & 34.1$\pm$7.5 & 33.2$\pm$8.2 & 33.5$\pm$8.0 & \textbf{33.7$\pm$7.7} \\
            & 64 & 40.4$\pm$7.2 & 41.8$\pm$7.7 & 39.9$\pm$7.8 & 40.1$\pm$7.8 & \textbf{40.6$\pm$7.7} \\
        \midrule
        \multirow{3}[2]{*}{\shortstack{RT-GENE \cite{FischerECCV2018}}} 
            & 16 & 46.7$\pm$12.7 & 43.8$\pm$12.2 & 46.8$\pm$13.6 & 54.5$\pm$11.9 & \textbf{48.0$\pm$13.2} \\
            & 32 & 54.4$\pm$12.6 & 53.6$\pm$12.2 & 53.4$\pm$13.5 & 55.7$\pm$11.6 & \textbf{54.3$\pm$12.5} \\
            & 64 & 57.6$\pm$12.5 & 58.8$\pm$11.6 & 55.7$\pm$13.3 & 56.1$\pm$11.5 & \textbf{57.0$\pm$12.3} \\
        \midrule
        \multirow{3}[2]{*}{\shortstack{RT-GENE \cite{FischerECCV2018}\\(Augmented)}} 
            & 16 & 24.6$\pm$10.1 & 28.6$\pm$13.2 & 24.7$\pm$11.6 & 20.1$\pm$10.8 & \textbf{24.5$\pm$11.9} \\
            & 32 & 29.4$\pm$10.9 & 36.4$\pm$15.5 & 29.1$\pm$12.0 & 23.1$\pm$11.2 & \textbf{29.5$\pm$13.4} \\
            & 64 & 37.2$\pm$12.4 & 43.4$\pm$16.1 & 34.8$\pm$12.7 & 26.4$\pm$11.7 & \textbf{35.5$\pm$14.7} \\
        \bottomrule
    \end{tabular}}}
\end{table}

\begin{figure}[t]
   \centering
   \subfigure[MPIIFaceGaze+Full-Face\cite{zhang2017s}]{
      \begin{minipage}[b]{0.45\linewidth}
         \includegraphics[width=1.0\linewidth]{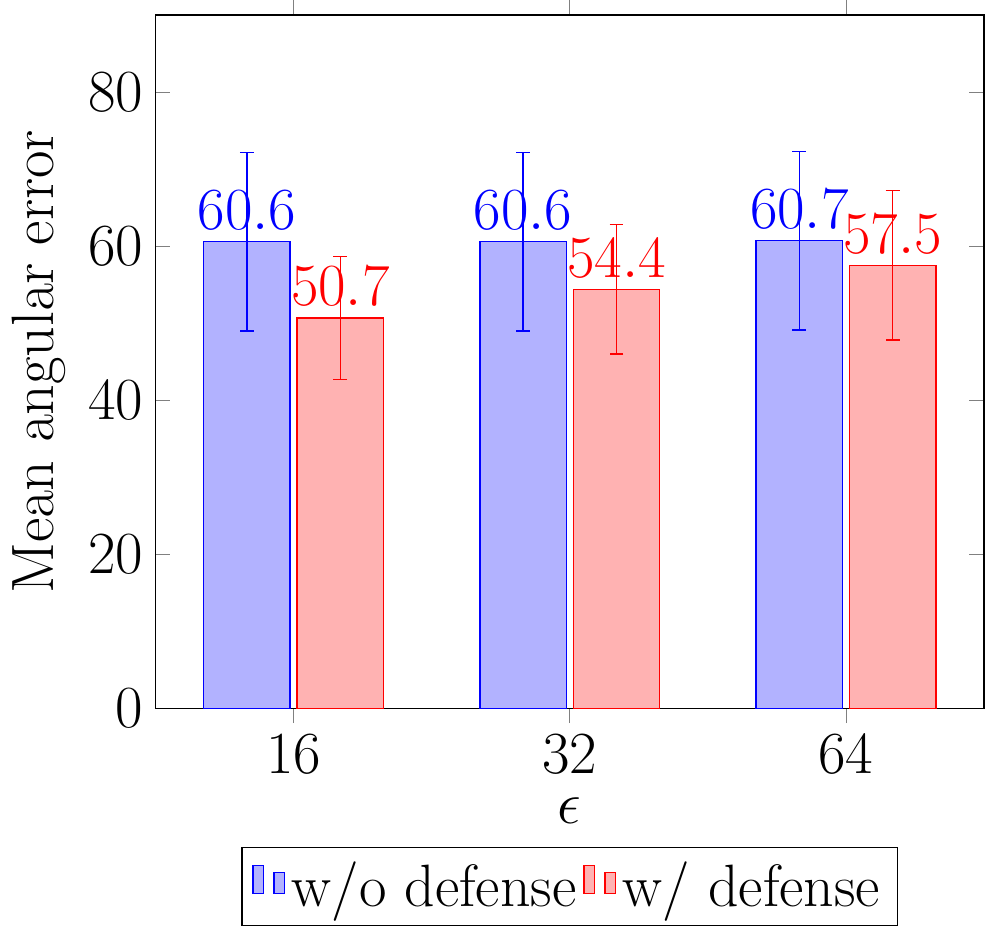}
      \end{minipage}
   }
   \subfigure[MPIIGaze\cite{DBLP:journals/corr/abs-1711-09017}+Gaze-Net\cite{DBLP:journals/corr/abs-1711-09017}]{
      \begin{minipage}[b]{0.45\linewidth}
         \includegraphics[width=1.0\linewidth]{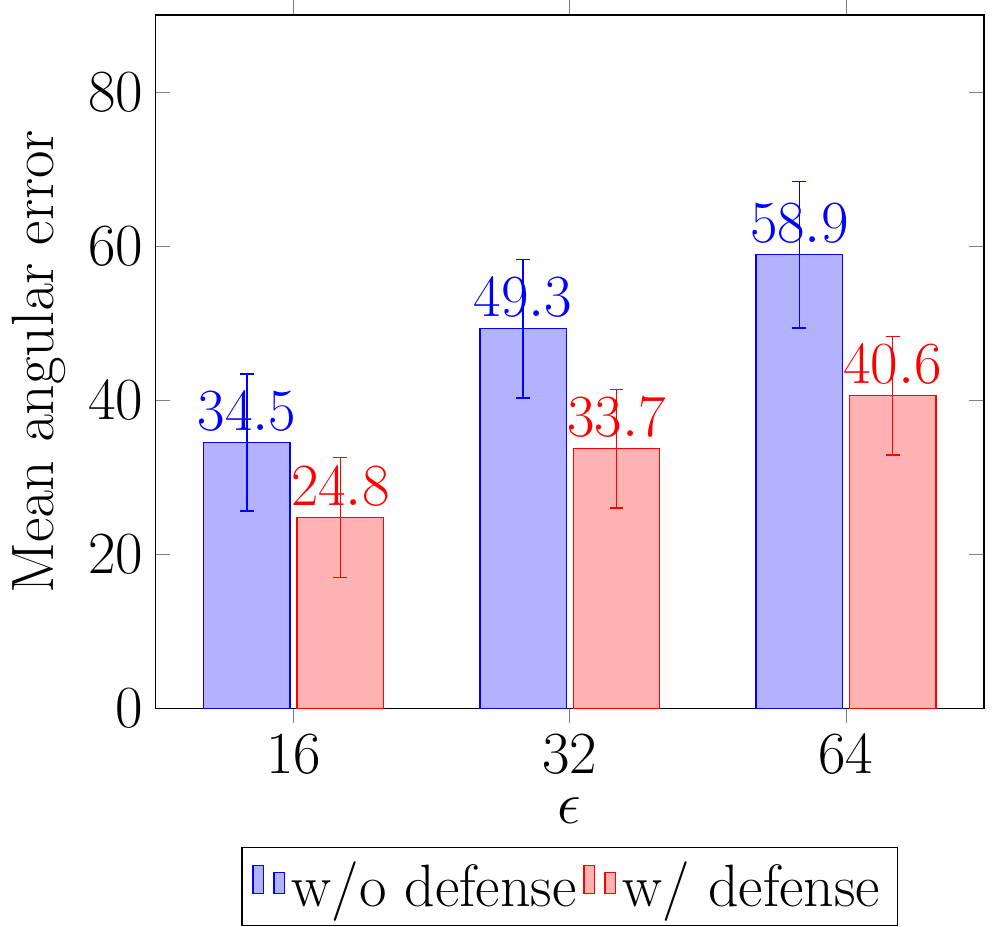}
      \end{minipage}
   }
   \subfigure[RT-GENE \cite{FischerECCV2018}]{
      \begin{minipage}[b]{0.45\linewidth}
         \includegraphics[width=1.0\linewidth]{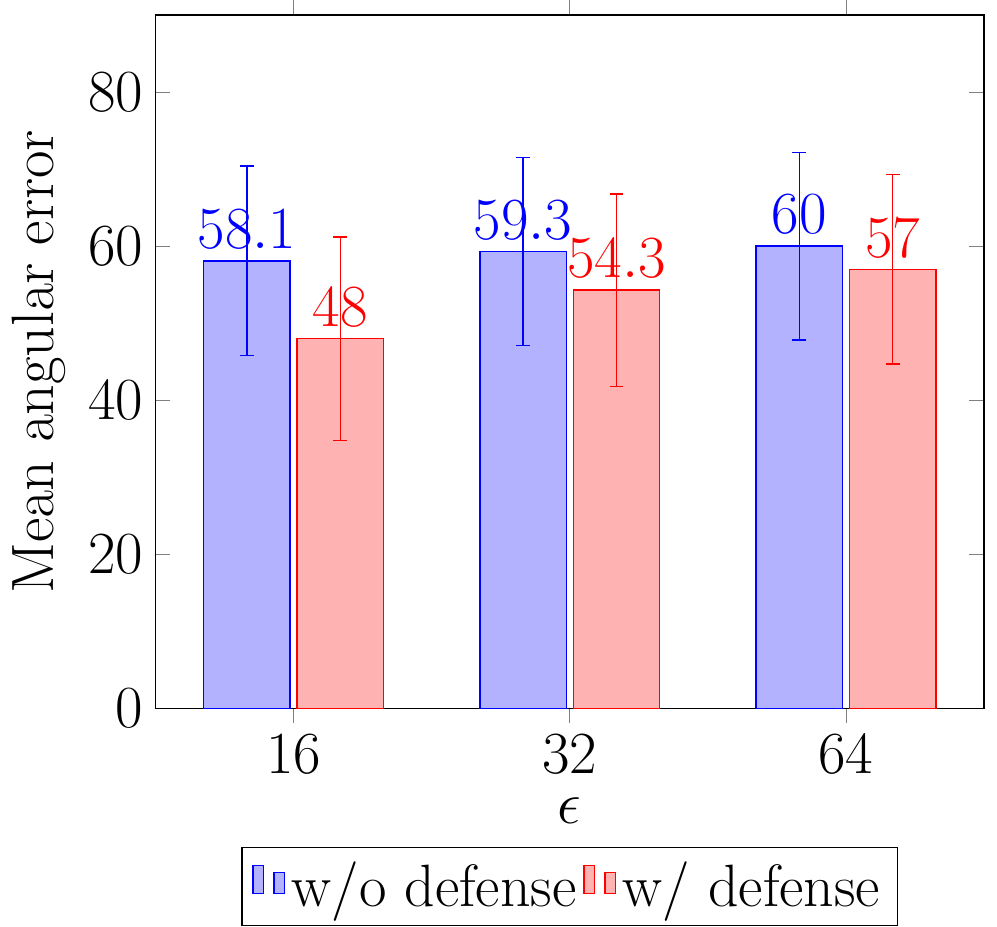}
      \end{minipage}
   }
   \subfigure[RT-GENE \cite{FischerECCV2018} (Augmented)]{
      \begin{minipage}[b]{0.45\linewidth}
         \includegraphics[width=1.0\linewidth]{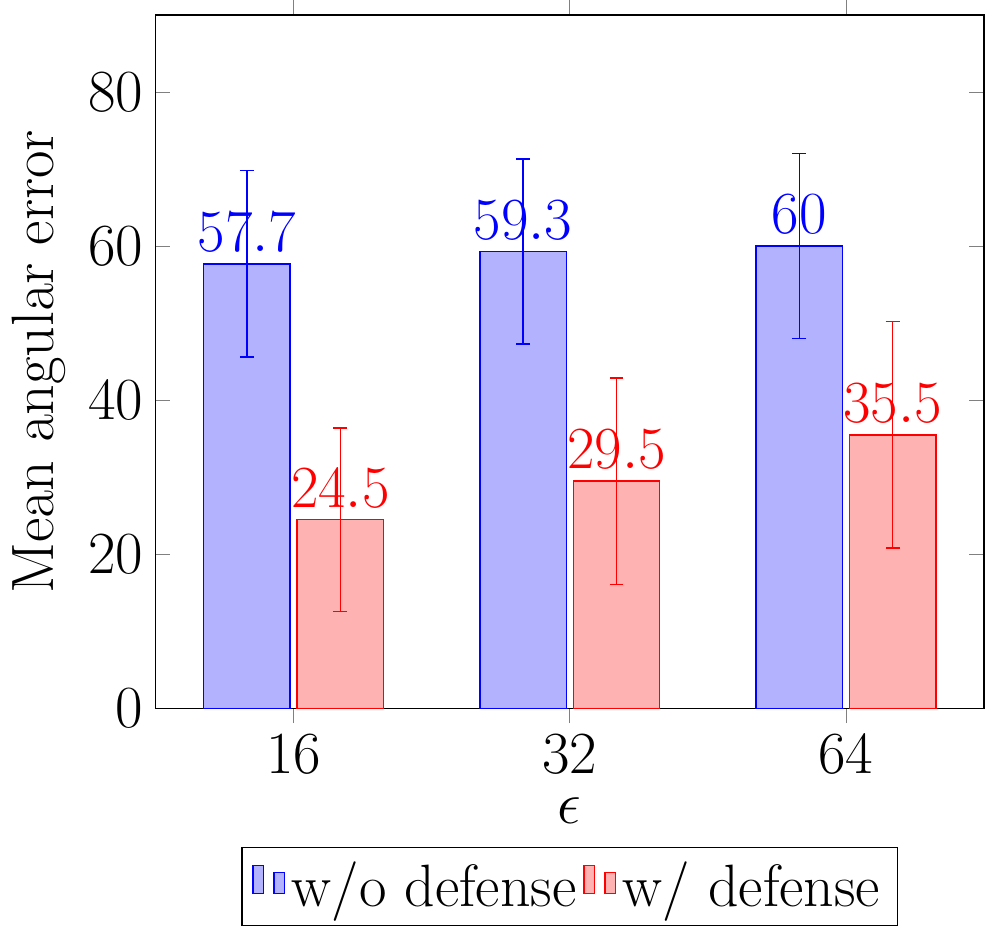}
      \end{minipage}
   }
   \caption{The mean angular error between the ground truth and the estimated gaze direction after attack without and with defense. The mean angular error of all the targets and all the inputs in one fold is displayed.}
\label{fig:defense}
\end{figure}


But knowing the angular error between the target and estimated gaze direction is not enough, because we don't know if estimated gaze direction after attack with defense becomes farther from or closer to the ground truth. Observed from Table \ref{table:gt_before}, \ref{table:gt_after} and Figure \ref{fig:defense} that all the models can make the estimated gaze direction after attack with defense becomes closer to the ground truth, which can be concluded that this defense method is effective for the defense of adversarial attacks on appearance-based gaze estimation tasks.

From these results, we conclude that for many appearance-based gaze estimation methods, we can use such defense method to reduce the vulnerability of models. 

\section{Conclusion}

In this paper, we have successfully shown the existence of the vulnerability of appearance-based gaze-estimation tasks. We studied the vulnerability in terms of attack and defense. On the attack side, we investigate the vulnerability from the aspects of parameter settings, universality, different face parts, attention patterns and smoothness. On the defense side, we demonstrate that the defense method can be used to reduce the vulnerability of models. This work draws the attention of the researchers to consider the vulnerability when designing appearance-based gaze estimation algorithms.


{\small
\bibliographystyle{ieee_fullname}
\bibliography{egbib}
}

\end{document}